%% file: root.tex
\documentclass[preprint,12pt]{elsarticle}

\usepackage{amssymb}
\usepackage{booktabs,array}
\usepackage{amsmath}
\usepackage{subcaption}
\usepackage{color,soul}
\usepackage{url}
\usepackage{listings}

\usepackage{siunitx}

\journal{Robotics and Autonomous Systems}

\begin{document}

\def\RobotName{MARLIN}

\begin{frontmatter}



\title{\RobotName: A Cloud Integrated Robotic Solution to Support Intralogistics in Retail}


\author[dfki]{Dennis Mronga}
\author[dfki]{Andreas Bresser}
\author[dfki]{Fabian Maas}
\author[dfki]{Adrian Danzglock}
\author[iai]{Simon Stelter}
\author[iai]{Alina Hawkin}
\author[iai]{Hoang Giang Nguyen}
\author[iai]{Michael Beetz}
\author[dfki,uni]{Frank Kirchner}

\affiliation[dfki]{organization={Robotics Innovation Center of German Research Center for Artificial Intelligence GmbH (DFKI)},
            addressline={Robert-Hooke-Str. 1}, 
            city={Bremen},
            postcode={28359}, 
            state={Bremen},
            country={Germany}}

\affiliation[uni]{organization={Robotics group of the University of Bremen},
            addressline={Robert-Hooke-Str. 1}, 
            city={Bremen},
            postcode={28359}, 
            state={Bremen},
            country={Germany}}

\affiliation[iai]{organization={Institute for Artificial Intelligence of the University of Bremen},
            addressline={Am Fallturm 1}, 
            city={Bremen},
            postcode={28359}, 
            state={Bremen},
            country={Germany}}

\begin{abstract}
In this paper, we present the service robot MARLIN and its integration with the K4R platform\footnote{K4R - Knowledge4Retail (\url{https://knowledge4retail.org})}, a cloud system for complex AI applications in retail. At its core, this platform contains so-called semantic digital twins, a semantically annotated representation of the retail store. MARLIN continuously exchanges data with the K4R platform, improving the robot's capabilities in perception, autonomous navigation, and task planning. We exploit these capabilities in a retail intralogistics scenario, specifically by assisting store employees in stocking shelves. We demonstrate that MARLIN is able to update the digital representation of the retail store by detecting and classifying obstacles, autonomously planning and executing replenishment missions, adapting to unforeseen changes in the environment, and interacting with store employees. Experiments are conducted in simulation, in a laboratory environment, and in a real store. We also describe and evaluate a novel algorithm for autonomous navigation of articulated tractor-trailer systems. The algorithm outperforms the manufacturer's proprietary navigation approach and improves MARLIN's  navigation capabilities in confined spaces. 
\end{abstract}



\begin{keyword}
Service Robotics \sep Digital Twin \sep Retail \sep Task Planning \sep Knowledge Processing


\end{keyword}

\end{frontmatter}


\section{Motivation}
\label{sec:introduction}
\input{sections/motivation.tex}

\section{Related Work}
\label{sec:related_work}
\input{sections/related_work.tex}

\section{\RobotName: A Service Robot to Support Intralogistics in Retail Stores}
\label{sec:service_robot}
\input{sections/service_robot.tex}

\section{The K4R Platform for AI Applications in Retail}
\label{sec:digital_twin}
\input{sections/digital_twin.tex}

\section{Experimental Evaluation}
\label{sec:evaluation}
\input{sections/evaluation.tex}

\section{Conclusion and Outlook}
\label{sec:conclusion}
\input{sections/conclusion.tex}


\section*{Acknowledgement}

This work has been performed in the Knowledge4Retail (K4R) project, funded by the German	Federal Ministry for Economic Affairs and Climate Action (grant number 01MK20001B).


\bibliographystyle{elsarticle-num-names} 
\bibliography{ref}





\end{document}

%% file: sections/motivation.tex
In order to be competitive with respect to international online sellers, the stationary retailer has  to combine its competencies in customer service and confidence with the possibilities of digitalization and robotics. In a future retail store, employees are providing advice to the customers, while robotic systems are in charge of stock-taking, replenishment and collecting scattered items. Smartphone apps direct the customer to the desired goods and answer their queries related to the assortment. Finally, the technology supports visually impaired or disabled people with shopping. Thus, AI and robotics in stationary retail have the potential to increase productivity by automating everyday tasks and improve customer experience. In order to put this vision into practice, detailed and comprehensive models of the store, the selling process and operation sequences need to be made available in machine readable form and provided to the robotic systems. The robots may benefit from this background information and improve their capabilities in terms of execution speed, autonomy, and safety. 

\begin{figure}[t]
\centering
\begin{subfigure}[c]{0.44\columnwidth}
\centering
\includegraphics[height=5.2cm]{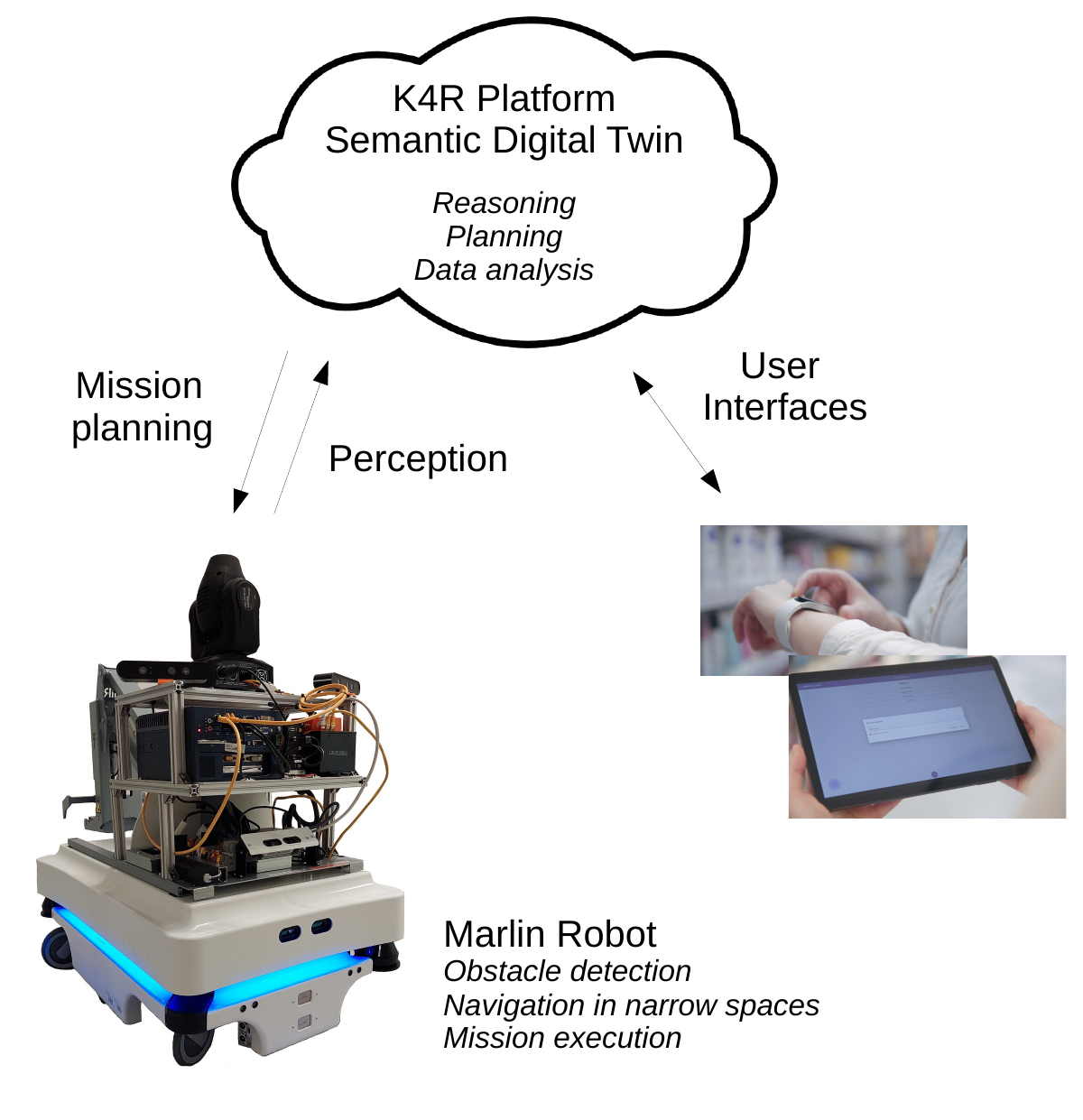}
\caption{System overview}
\label{fig:overview}
\end{subfigure}
\hfill
\begin{subfigure}[c]{0.55\columnwidth}
\centering
\includegraphics[height=5.2cm]{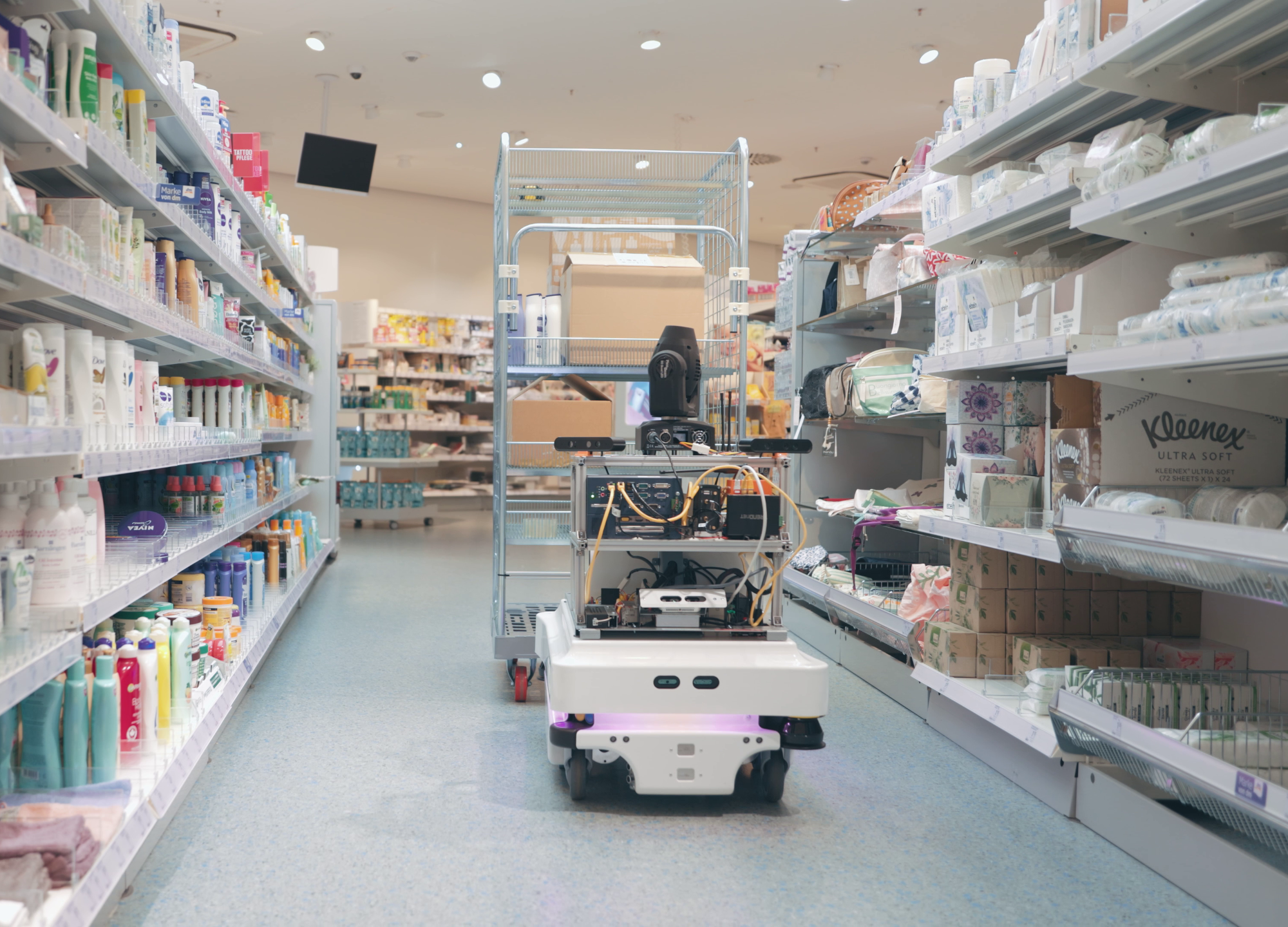}
\caption{Application: Support of store employees in retail}
\label{fig:application}
\end{subfigure}

\end{figure}

In this paper, we describe the mobile service robot \RobotName{} (Mobile Autonomous Robot for intra-Logistics IN retail) and its integration with the K4R platform, an open-source cloud platform for AI and robotic applications in retail (see Figure~\ref{fig:overview} for an illustrative overview). At its core, this platform provides so-called semantic digital twins, a generic, machine-readable format for digital representation of retail stores. They allow the construction of realistic digital worlds and enable a variety of novel AI and robotics applications like data analysis, symbolic reasoning, and process planning. We exploit the potential of integrating \RobotName{} with the K4R platform and evaluate its capabilities in terms of perception, autonomous navigation, and task planning. The overall system is evaluated in a retail intralogistics scenario, namely the support of store employees in refilling shelves (see Figure~\ref{fig:application}). The robot autonomously transports goods to the target shelf, assists employees with replenishment using a pointer unit, and interacts with them via a graphical interface. At all times, it exchanges information with the semantic digital twin, for example the position of products within the shelves, the whereabouts of store employees, as well as the location of obstacles in the corridors, which are perceived through 3D onboard sensors on the robot. This way \RobotName{} can safely navigate in a retail store and adapt to unforeseen changes in the environment, e.g., crowded, impassable aisles. 

The robot itself consists of a commercially available MiR100 platform with a mounted transport hook and additional equipment such as sensors and a pointer unit. The hook can be used to attach and transport trolleys, which are typically used in retail when stocking shelves. During transport of trolleys the robot represents a tractor-trailer system with  variable and comparatively large footprint. To reliably navigate within the confined environment of a retail store, sophisticated navigation planning and execution skills are required for such a system. Thus, we also develop and evaluate an approach for autonomous navigation of articulated tractor-trailer systems, which overcomes certain limitations of classical navigation methods in confined spaces. 

In summary, the contributions of this paper are:

\begin{itemize}
\item The introduction of \RobotName{}, a mobile robotic solution to support shelf intralogistics in retail stores
\item A simple, yet efficient approach for detecting and classifying obstacles in a retail environment based on 3D sensors
\item The application of a semantically annotated digital twin that stores information about the retail store and allows semantic queries
\item An algorithm for autonomous navigation of articulated tractor-trailer systems, which outperforms the robot manufacturer's navigation approach
\end{itemize}

The paper is structured as follows. In Section~\ref{sec:related_work}, we relate our work to the state of the art in robotics for retail and intralogistics. In Section~\ref{sec:service_robot}, we describe the service robot \RobotName{} and its capabilities in terms of perception, and autonomous navigation.  In Section~\ref{sec:digital_twin}, we first describe the semantic digital twin and its connection to \RobotName{}, followed by an explanation of our approach to autonomous task planning, as an example of the use of AI applications in the K4R platform. In Section~\ref{sec:evaluation}, we present experimental results on obstacle classification, autonomous navigation of tractor-trailer systems, as well as task planning utilizing the semantic digital twin. Finally, Section~\ref{sec:conclusion} provides a short conclusion and outlook on future applications.

%% file: sections/related_work.tex
This section provides a state-of-the-art overview on the different areas of research touched by this work, namely robotics for retail and intralogistics applications in general, the digital twin technology, autonomous navigation for tractor-trailer systems, as well as adaptive task planning.

\subsection{Robotics in Retail}

When related to the field of warehouse logistics the number of commercially available robotic systems for stationary retail is comparatively low. One reason is the greater complexity of the store environment and the associated challenges in terms of robotic perception, navigation, and manipulation. For example, the store might be filled with customers, which impede an autonomous robot from navigating efficiently and safely. Furthermore, the products on the shelves vary greatly in terms of size, shape and weight, which makes autonomous manipulation difficult. Finally, the stores in itself vary widely and a one-fits-all robotic solution does not exist. Therefore, only a few autonomous robots have been used efficiently and economically in stationary retail to date, although the growth is rapid, as mentioned by~\citet{Bogue2019}.  

Positive examples of economically viable robotic applications in stationary retail exist in the area of inventory and out-of-stock detection, for example the Tally robot by simbe robotics~\cite{tallyRobot2022}, or the AdvanRobot system proposed by~\citet{Morenza-Cinos2017}. In the area of inventory and out-of-stock detection, the visual perception of articles is in focus. A survey of machine vision based retail product recognition systems is presented by~\citet{Santra2019}. The work of~\citet{Kumar2014} describes semi-automatic out-of-stock detection using mobile robots and virtual reality. The approach is evaluated in a mock retail store. The same application is targeted by~\citet{Paolanti2017}. The authors use deep convolutional neural networks to automatically detect out-of-stock events in a real store environment during working hours. In the REFILLS project\footnote{\url{http://www.refills-project.eu}} a mobile autonomous robot is presented to acquire models of retail stores, count the stocked products, and document their arrangements in the shelves~\cite{beetz2022robots}. 

\subsection{Digital Twins}

Digital twins are increasingly used in industrial manufacturing and production to represent, simulate, monitor, analyze, and optimize production processes and product lifetime cycles~\cite{Singh2021}. However, in the area of stationary retail, the use of digital twins is less widespread. The digitization of stationary retail demands for an integration of various, disparate information like product data, article localization, customer routes and attention, as well as sales figures.

A considerable challenge for autonomous robots in retail is the perception and interpretation of the store environment (not the individual products in the shelves), which is different from store to store, but might also change within same store on a daily basis due to varying placement of articles, stock level, or locations of stand-up displays. A digital twin of the store environment can be used to collect, preprocess, and provide the perceived data from multiple sources in a machine-readable format. However, establishing a digital environment from scratch without considerable manual effort is a complex problem.  \citet{Paolanti2019} describe a semantic object mapping based on 3D point cloud data to analyze and map a store environment. The work by~\citet{beetz2022robots} introduces semantic store maps, which are a special form of \textit{semantic digital twins (semDT)} as described by~\citet{Kuempel2021}. A semDT is a semantically enhanced virtual representation of the physical retail store, which connects symbolic knowledge with a scene graph, allowing complex reasoning tasks. The work presented in this paper builds upon the concept of the semDT and showcases a robotic application to support store employees in shelf refilling. Specifically, we use the reasoning capabilities of KnowRob~\cite{beetz2018know}, a knowledge processing system for robots. KnowRob organizes the digitized store data coming from disparate sensor sources and allows a robotic task planner to pose semantic queries on this data.


\subsection{Autonomous Navigation of Tractor-Trailer Systems}

Autonomous navigation for mobile robots is a widely researched field and there are many commercially available solutions, especially for indoor environments~\cite{mir2022,clearpathrobotics2022}. The application to tractor-trailer systems, i.e., an actuated towing vehicle (tractor) with attached passive trailer, is more challenging in terms of navigation planning and path following. Application areas for autonomous tractor-trailer systems include agriculture and autonomous driving in road traffic. 

In agriculture, automated harvesting in particular is an important issue. The work by~\citet{Thanpattranon2016} introduces a method for controlling an autonomous agricultural tractor equipped with a two-wheeled trailer using a single laser range finder. To simplify navigation in narrow rows of orchards and plantations, they developed a sliding mechanical coupling between trailer and tractor, which adjusts the position of the hitch-point. In contrast, we develop a software solution that can be applied to various tractor-trailer systems without the need to mechanically adjust the hitch point. The work by~\citet{BACKMAN201232} introduces a non-linear model predictive path tracking approach for agricultural machines. In contrast to the method presented by~\citet{Thanpattranon2016} the authors fuse information from GPS, laser range finder, and IMU to estimate tractor position and heading angle. However, they deal only with the problem of trajectory tracking, not with planning. Moreover, their mechanical structure includes an additional degree of freedom, namely a hydraulic joint in the suspension between the towing vehicle and the trailer. This leads to a multivariate control problem that is more complex than that of a passive trailer, although it offers more flexibility. 

In the field of autonomous driving in road traffic, trucks with trailers are of particular interest. The work by~\citet{Oliveira2020} introduces an optimization-based path planning algorithm for articulated vehicles. They formulate the on-road path planning as an Optimal Control Problem (OCP), which is solved using Sequential Quadratic Programming, thereby comparing different cost functions. Results are, however, presented only in simulation. Similarly,~\citet{Li2019} formulate path planning for tractor-trailer systems as an OCP. In their approach, N trailers may be attached to a single tractor system. An initial guess for the OCP is provided by a sample-and-search planner. Again, results are presented in simulation only and computation times are large (about 30-180 seconds depending on the planning problem). In contrast, the approach proposed by~\citet{Shen2021} allows real-time trajectory planning. The authors use a sampling based planner to plan ahead a set of trajectories. Afterwards, they perform collision checking by forward simulating the trajectories and selecting the one with the lowest cost, where the cost function is a combination of guidance, collision, smoothness and deviation costs. The authors evaluate their approach in simulation, considering three different road profiles with varying curvature. However, the evaluation scenarios are quite simple, and it is unclear whether the approach will retain its high performance in a more complex environment. 

In contrast to the approaches described above, our target environment is a retail store where narrow aisles and static as well as dynamic obstacles are common. Thus, we are faced with a complex planning problem, moderate requirements on computation time for global planning, as well as real-time requirements on local path planning and obstacle avoidance. Optimal control has some interesting properties, e.g. it allows the integration of dynamic constraints of the tractor-trailer system. However, it entails very long computation times. Therefore, it is out of question for our main use case, the support of retail intralogistics. The tractor-trailer system we consider consists of a MiR100 mobile robot with differential drive (tractor) and a trailer without actuation. The trailer is rigidly attached to a transport hook that has a passive rotational joint located on the main rotation axis of the mobile robot. This differs from most tractor-trailer systems, where the trailer joint is usually located on the rear axle of the tractor.

\subsection{Task Planning for Mobile Robots in Industry and Retail}

Task planning in robotics is concerned with deliberately deciding on a sequence of actions to take in order to achieve a given set of goals~\cite{siciliano2008springer}. AI planning methods have a long history~\cite{Fikes1972} and have been applied to increasingly complex robotic applications, for example household~\cite{Beetz2010} or manufacturing~\cite{Huckaby2013}. Despite the long exploration of AI-based planning there are still open research problems, e.g., how to efficiently represent knowledge from disparate sensor sources, or how to bridge the gap between symbolic and numerical action representations. Task planning for robotics is closely linked to the fields of knowledge representation and reasoning. Planning complex tasks in real-world environments requires a powerful representation of knowledge acquired from disparate sources, as well as a means to reason about this information. Both can be provided by KnowRob~\cite{beetz2018know}, the knowledge representation and reasoning framework which we use in our work. We connect KnowRob to the knowledge base of ROSPlan~\cite{Cashmore2015}, a framework for AI planning in robotics, which provides various planners. 

In industry and retail most robotic mission or task planning approaches are concerned with intralogistics, e.g., managing a fleet of AGVs and other agents to optimize warehouse logistics~\cite{Bolu2021,Shi2022}. Although the primary goal in this research is to plan autonomous transport tasks for a robot navigating a retail store, our planning approach allows arbitrary tasks like interacting with the store employees, manipulation of products, or updating the digital store representation. 

%% file: sections/service_robot.tex
This section describes the service robot \RobotName{} and its capabilities regarding perception, autonomous navigation and interaction with the store employees.

\subsection{System Description}

\begin{figure}
\centering
\includegraphics[height=6.7cm]{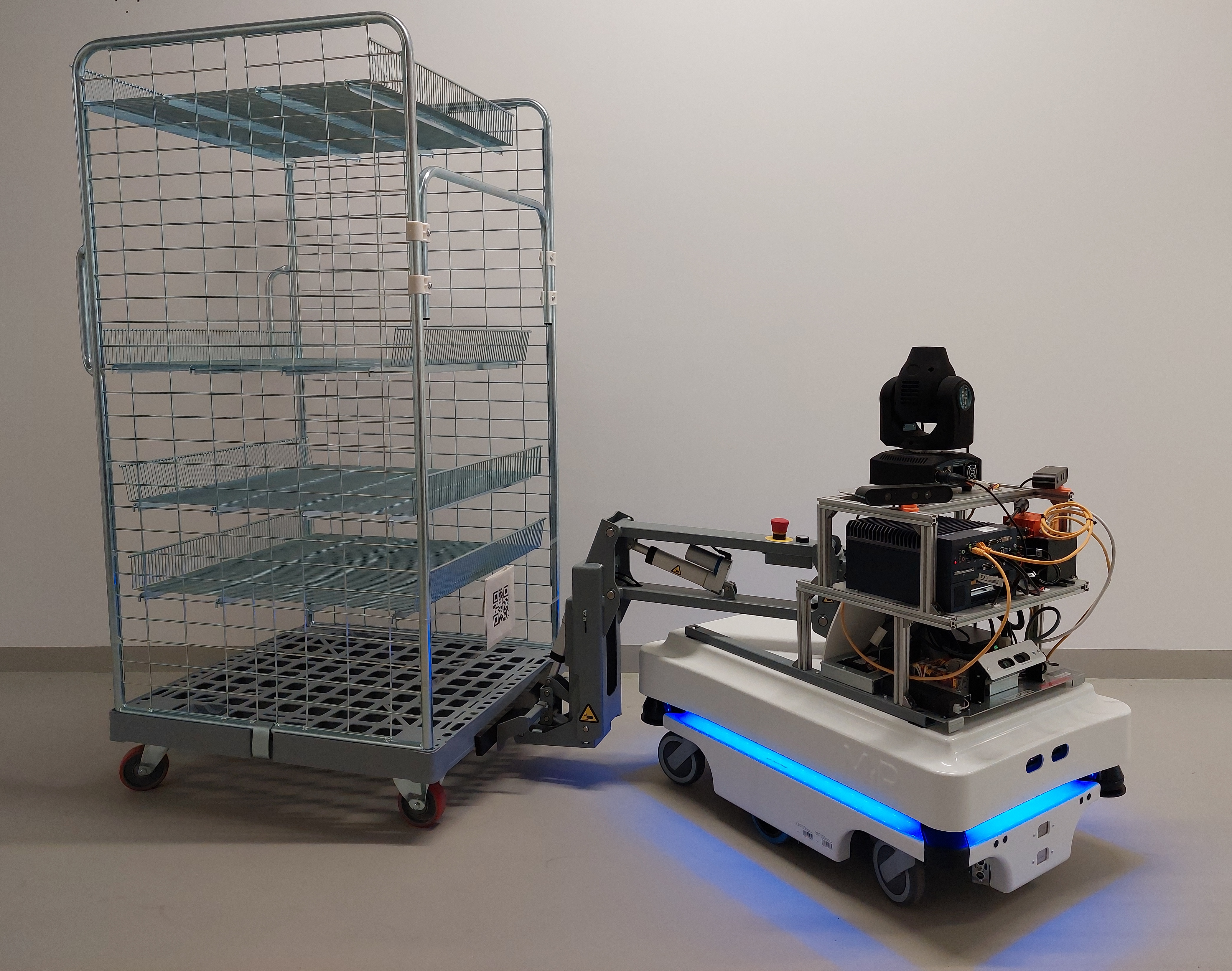}
\hfill
\includegraphics[height=6.7cm]{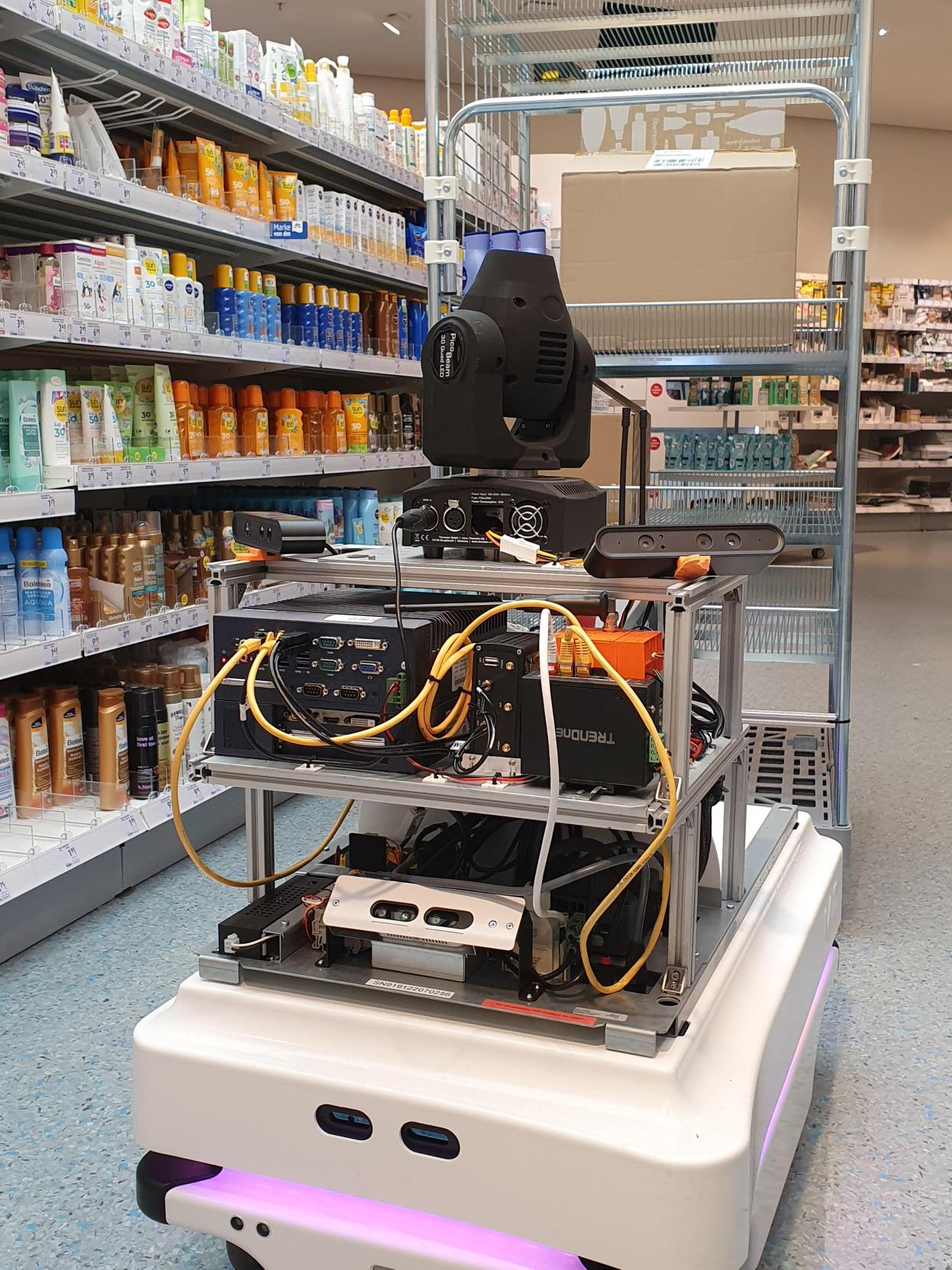}
\caption{\RobotName{}: A mobile service robot for the support of store employees}
\label{fig:mir100_hook}
\end{figure}

The design of \RobotName{} as illustrated in Figure~\ref{fig:mir100_hook} has been led by the requirements of pilot application \textit{Service Robotics to Support Store Employees} in the Knowledge4Retail (K4R) project. Within this project, a mobile, autonomous robot should be developed to support store employees in shelf refilling. The robot should be able to navigate efficiently and safely within a retail store. Apart from that it should (a) be integrated seamlessly with the K4R platform, a cloud solution to enable AI applications in stationary retail, (b) reuse existing structures of the stores, e.g., carts on wheels used for intra-logistics, and (c) provide user interfaces to interact with the store employees.

\RobotName{} consists of a commercially available MiR100 platform\footnote{\url{https://www.mobile-industrial-robots.com/solutions/robots/mir100/}} with transport hook, equipped with an external PC, a pointer unit to guide the store employee in the process of replenishment, as well as 4 RGB-D cameras, which provide points clouds in a 360 degree view. The system is able to autonomously pick-up, carry, and place transport carts on wheels, which are commonly used by retailers. Interaction with the user can be performed via an attached tablet, which is connected via the K4R platform as described in Section~\ref{sec:digital_twin}.

\subsection{Obstacle Detection and Classification}

\begin{figure}
\centering
\includegraphics[width=\columnwidth]{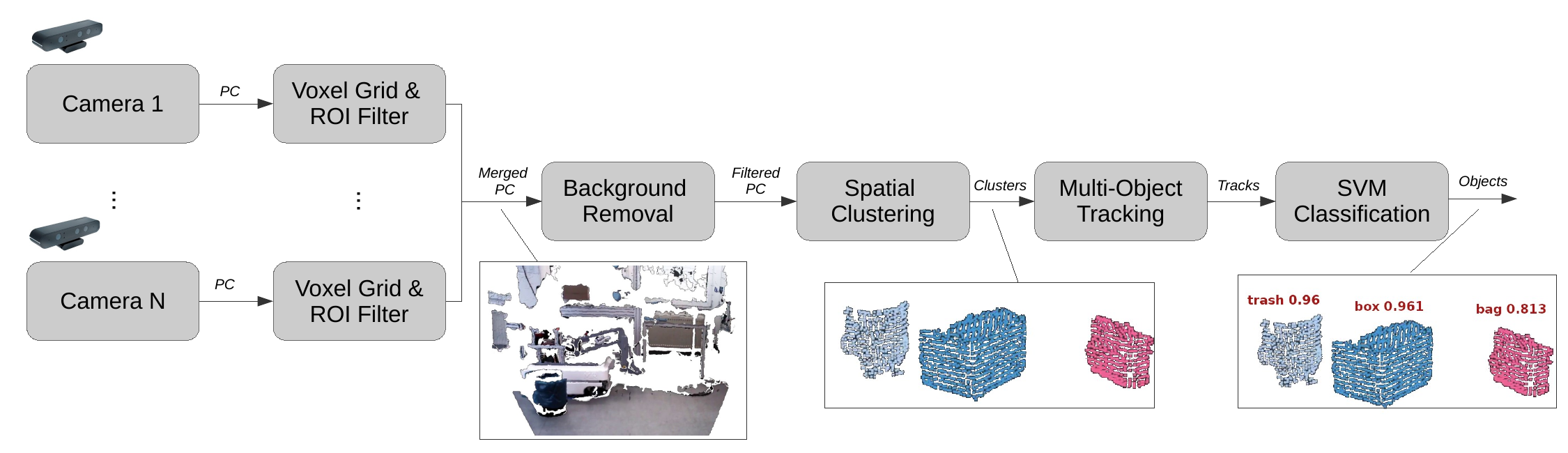}
\caption{Sensor Processing Pipeline for obstacle detection on \RobotName{}. PC - Point Cloud.}
\label{fig:sensor_processing_pipeline}
\end{figure}

By integrating MARLIN with the K4R platform, the robot continuously exchanges information with the retail store's semDT. Using its built-in sensors, the robot can detect obstacles, upload their position to the semDT, and reuse this information for future task planning and navigation. To handle static and dynamic obstacles, a pipeline was developed to detect and classify objects in the raw point cloud data. This pipeline, which is an extension of the multi-object tracking described by~\citet{DEGEAFERNANDEZ2017102}, contains three main processing steps: (1) background removal, (2) clustering and tracking of objects, and (3) normalization \& classification of tracked objects. Figure~\ref{fig:sensor_processing_pipeline} shows an overview of the sensor processing pipeline.

\begin{figure}
  \centering
  \includegraphics[height=10cm]{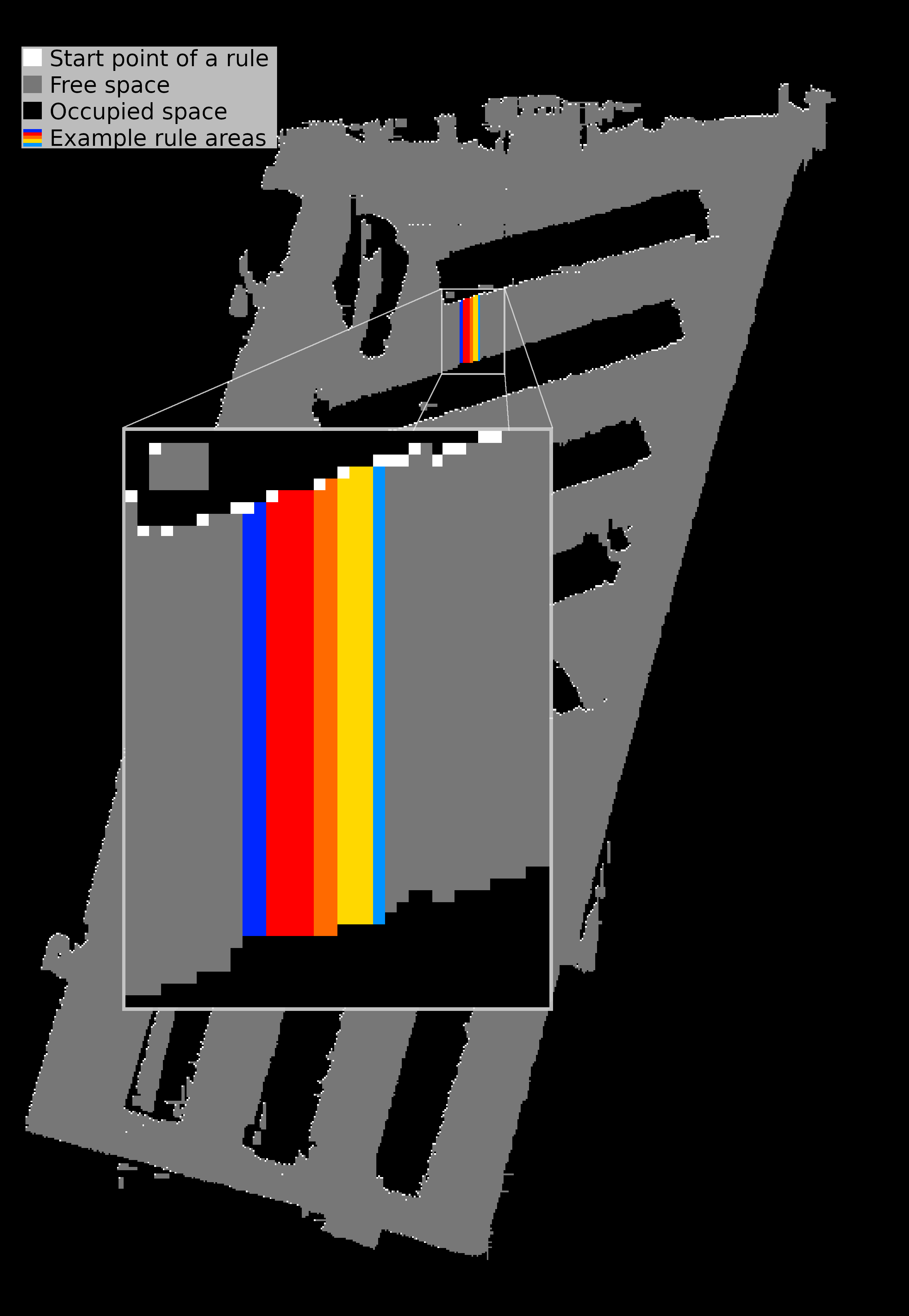}
  \caption{Visualization of the whitelist rules on a map of a retail Store}
  \label{fig:map_filter_1}
\end{figure}

\paragraph{Background Removal} The multi-object tracking approach we use to cluster and track obstacles in raw point cloud data has originally been developed for stationary robots~\cite{DEGEAFERNANDEZ2017102}. In the original approach, the stationary background is removed from the point cloud data before clustering in order to increase performance. While this task is trivial for a stationary system, in a mobile robotics application the background filter must constantly adapt to the environment and be much faster to ensure that no artifacts remain in the filtered point cloud, even when the robot is moving fast. Thus, we use the following procedure for background removal. We first reduce the number of points in the original point cloud using a voxel grid filter. Points that are too far away or too close to the ground are also removed. Then, the point clouds from each camera are merged into a single point cloud and irrelevant areas are removed using a rule-based filter implemented in the Point Cloud Library~\cite{Rusu2011}. This filter removes all 3D points corresponding to shelves and walls of the retail store using whitelist rules. The whitelist rules, defined by rectangles in the 2D map, are determined as follows: We iterate row-wise through the pixels of the 2D map until we find a free pixel. This defines the upper left corner of the first rectangle. Then we move in x- and then in y-direction until we hit a occupied pixels, which provides the bottom right corner of the rectangle. We repeat this procedure, until all the free pixels in the map are covered by rectangles. The filter then removes all 3D points whose xy coordinates are not covered by a whitelist rule. Figure~\ref{fig:map_filter_1} shows a subset of the rules defined for the 2D map of a retail store. The starting points of each rule are shown in white and the areas to be removed are shown in black. The enlarged image shows the whitelist rules, each in a different color. By using the rule-based filter, shelves and other permanent (known) obstacles in the store are filtered out of the point cloud, while dynamic obstacles remain.

\begin{figure}
  \centering
  \begin{subfigure}[b]{0.49\textwidth}
         \centering
         \includegraphics[height=5cm]{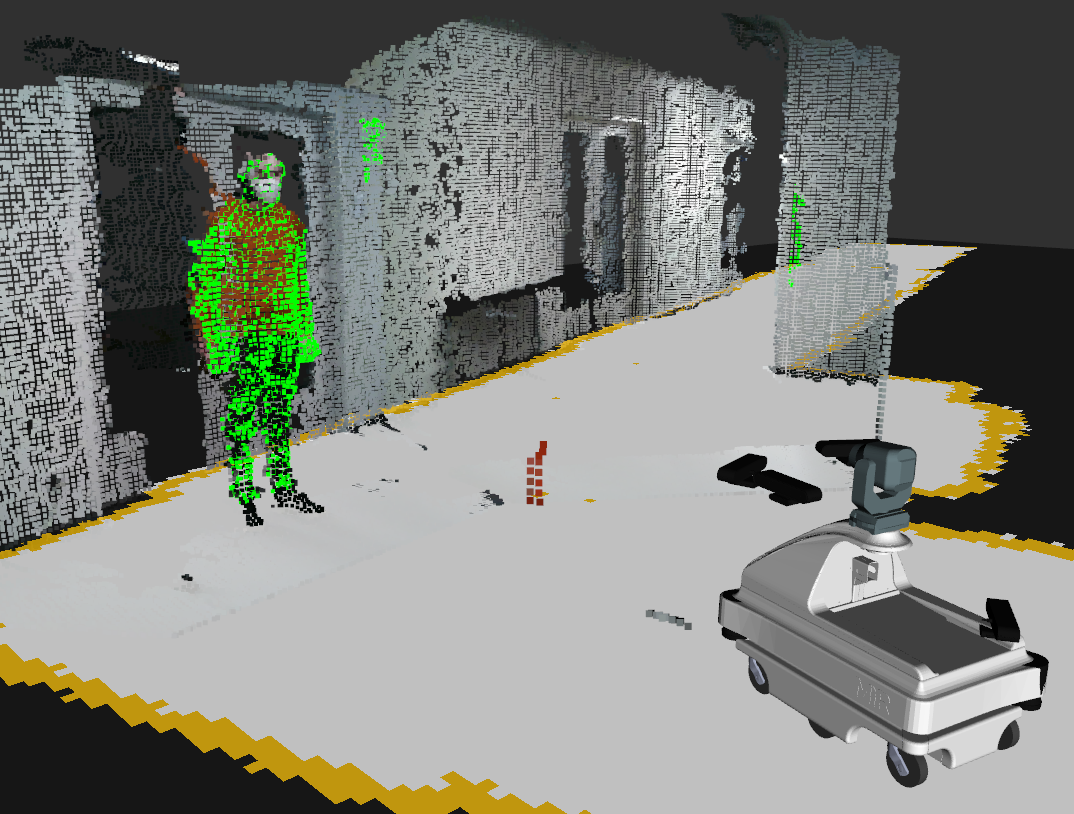}
         \caption{Pointcloud without (normal color) and with filter(green dots) \newline}
         \label{fig:pointcloud_filter_1}
  \end{subfigure}
  \begin{subfigure}[b]{0.49\textwidth}
         \centering
         \includegraphics[height=5cm]{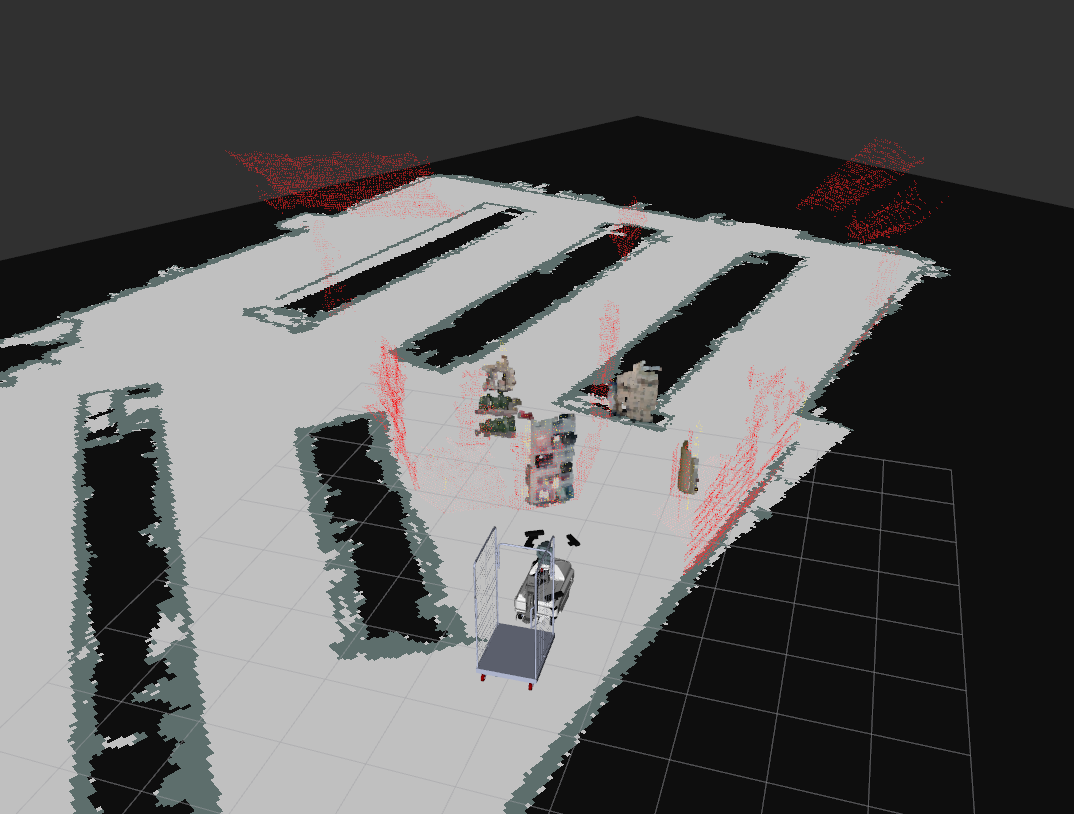}
         \caption{Filtered Pointcloud of an environment with multiple different obstacles (unfiltered - red, clustered points - original color)}
         \label{fig:pointcloud_filter_2}
  \end{subfigure}
  \caption{Results of the Pointcloud Filter.}
  \label{fig:pointcloud_filter}
\end{figure}

\paragraph{Clustering and Tracking}
The filtered point cloud is passed to an adapted version of the multi-object tracking described by~\citet{DEGEAFERNANDEZ2017102}. In this approach, the filtered 3D points are clustered according to their Euclidean distance. Small clusters are removed as they are assumed to result from sensor noise. In the subsequent object tracking, the remaining clusters are modeled as 3D ellipsoids, which can be used to estimate the full pose and spatial velocity of each cluster. For tracking, a track ID is assigned to each cluster. Given a set of new point clouds from the processing pipeline and a set of tracks, the tracking algorithm computes an updated set of tracks as follows. First, the states (poses and spatial velocities) of all tracks are updated using a Kalman filter. Then, each existing track is assigned to a cluster using an association measure based on the Euclidean distances of the 3D points in the cluster to the ellipsoid of the track. In this way, objects are tracked over multiple time frames. The method is robust to partial occlusion and sensor noise and allows tracking of multiple objects in cluttered scenes.

\begin{figure}
\centering
\begin{subfigure}{0.32\columnwidth}
\includegraphics[width=\columnwidth]{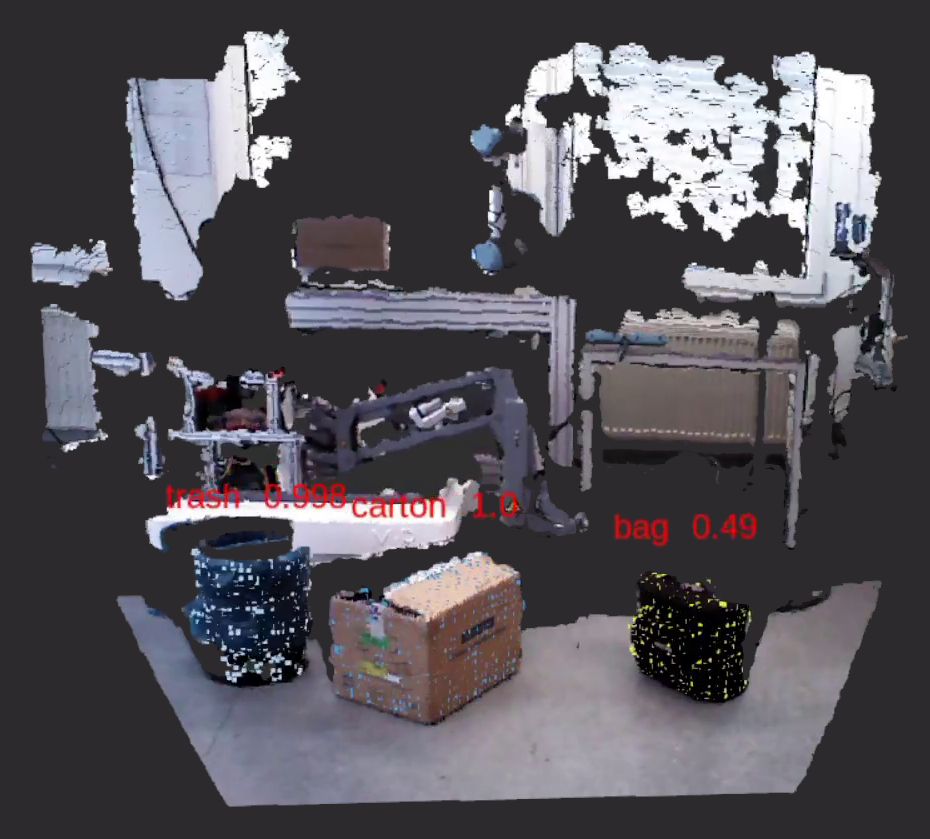}
\caption{Original 3D scene}
\label{fig:sensor_proc_example_1}
\end{subfigure}
\hfill
\begin{subfigure}{0.32\columnwidth}
\includegraphics[width=\columnwidth]{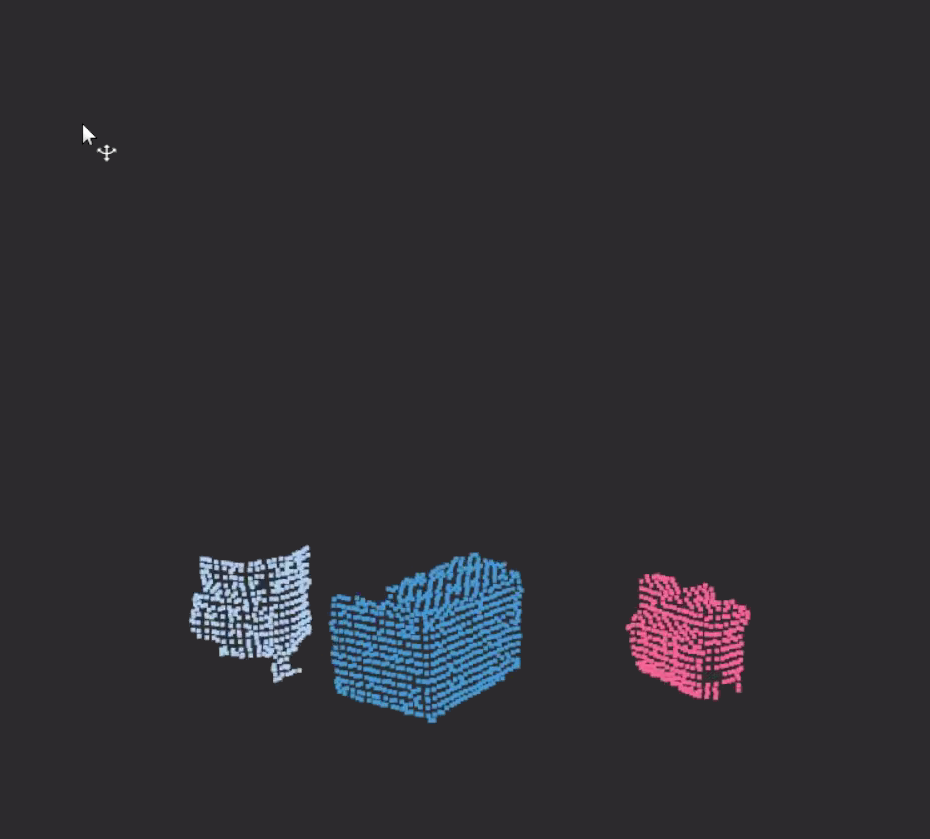}
\caption{Segmented clusters}
\label{fig:sensor_proc_example_2}
\end{subfigure}
\hfill
\begin{subfigure}{0.32\columnwidth}
\includegraphics[width=\columnwidth]{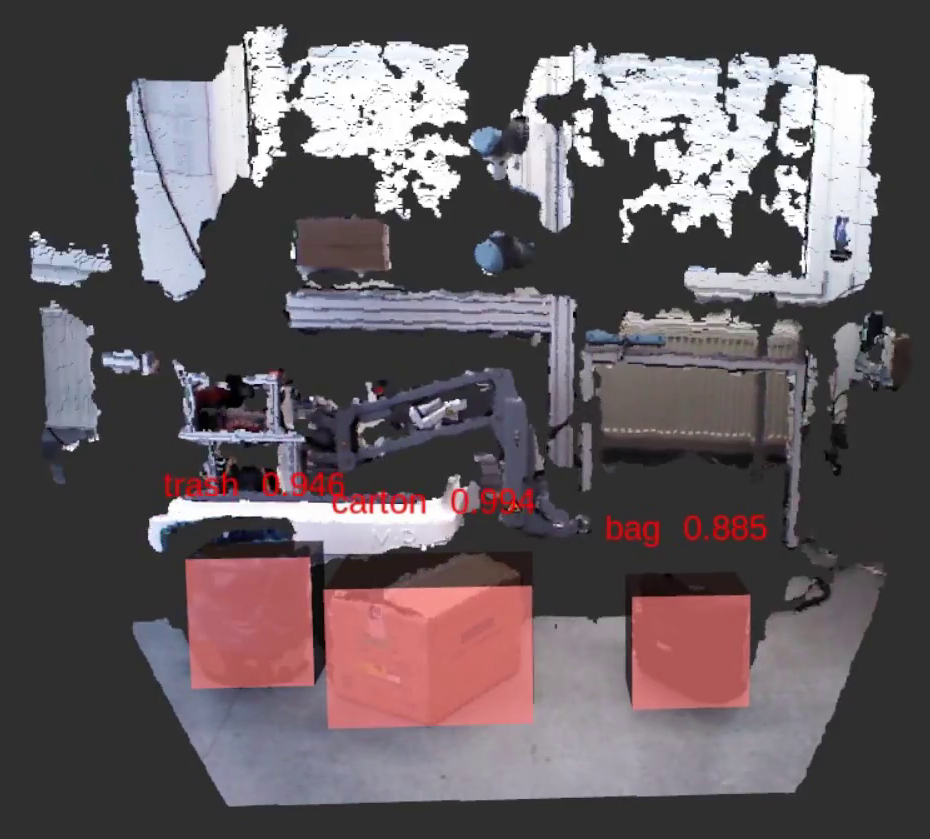}
\caption{Result after classification}
\label{fig:sensor_proc_example_3}
\end{subfigure}
\caption{Clustering, tracking and classification of three different objects in a laboratory environment.}
\label{fig:sensor_proc_example}
\end{figure}

\paragraph{Normalization and Classification} We use obstacle classification based on 3D point clusters provided by the multi-object tracking. The idea is that permanent obstacles, e.g., display stands, can be added to the virtual environment in semDT to improve the autonomous navigation of the mobile service robot, while non-permanent obstacles such as customers or shopping carts can be ignored. The clusters are normalized to have zero mean and unit variance. For each cluster the surface normals are computed\footnote{\url{https://pcl.readthedocs.io/en/latest/normal_estimation.html}} and then forwarded to a prediction node as one-dimensional feature vectors, which have a fixed size $n=10000$. If a cluster provides less features, the feature vector is artificially augmented. The prediction node computes the probability that a given cluster belongs to a given object class. We compare different classification approaches such as random forests (RF), support vector classifier (SVC), Gaussian processes (GP), a voting classifier (VC) consisting of a random forest and a support vector classifier, and stochastic gradient descent (SGD). We chose SVC above other approaches~\cite{mousavian20173d,qi2017pointnet,zhou2017voxelnet}, because it is a simple yet computationally efficient method and produces decent results with the recorded training dataset. The model was trained with a linear kernel. All implementations were from the Scikit Learn~\cite{scikit-learn2011} library. To reduce computation time, multiple clusters are evaluated simultaneously. The models of the classifiers are trained offline using a relatively small dataset consisting of raw point clouds of each object class. Figure~\ref{fig:sensor_proc_example} shows initial results in a laboratory environment, including the original scene (left), background filtering and clustering (center), and classification (right) using the five different object classes.
 
\subsection{Autonomous Navigation for Tractor-Trailer Systems}

The proprietary navigation stack of the MiR100 platform provides navigation capabilities for indoor operation through a ROS (Robot Operating System) interface~\cite{Eppstein2010}, both for operation with and without an attached trailer. Autonomous navigation with an attached trailer, however, requires enormous safety distances around the footprint of the robot (the recommended minimum corridor width is the total length of the system plus $50\mathrm{\hspace{2pt}cm}$ safety distance). Therefore, the system is not able to navigate through narrow aisles typically found in the defined target environment, a retail store. However, physically this is quite possible, as a human operator is able to remote control the system safely in the target environment. For this reason, we implement and integrate a novel approach for autonomous navigation of tractor-trailer systems, which provides improved capabilities compared to the proprietary approach.

%
%
%

\begin{figure}[t]
\centering
\includegraphics[height=150pt]{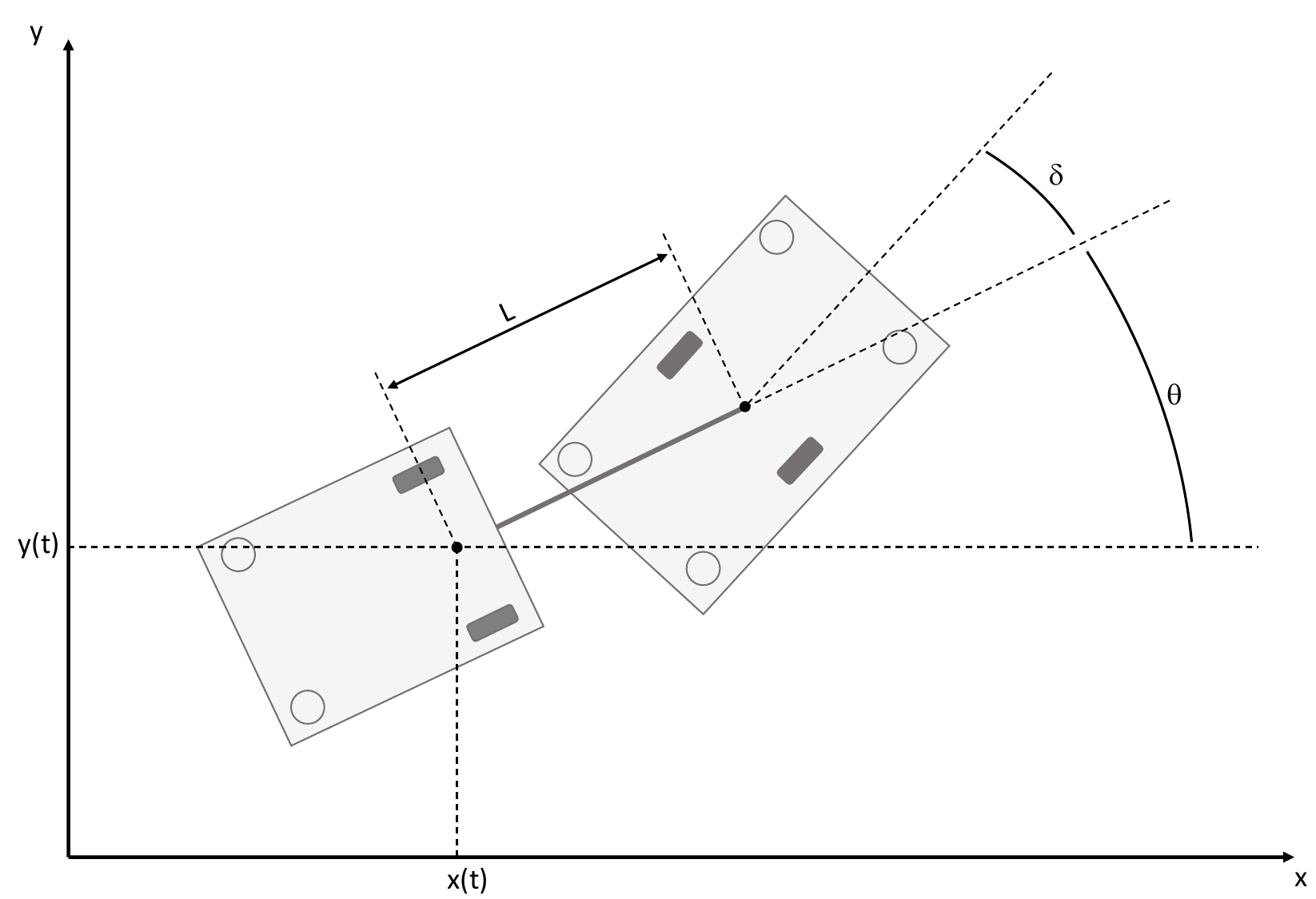}
\caption{Model of the tractor-trailer system kinematics according to a bicycle model.}
\label{fig:nav-kinematics}
\end{figure}

\paragraph{Vehicle Kinematics}
For the differential drive tractor with the trailer joint attached directly at the steering axis, a car-like controller with kinematic bicycle model can be applied, as shown in Figure~\ref{fig:nav-kinematics}. In our model, the trailer represents the rear axle of the model and the MiR100 represents the steering axle. The cart is attached to the robot via a transport hook, which has a passive rotational joint located at the rotation axis of the robot base. After picking up the cart, it is rigidly attached to the hook. 

The trailer has one axle with fixed caster wheels and one axle with swivel caster wheels that can passively rotate around their vertical axis to follow the motion of the trailer. We define a coordinate frame in the center of the axle between the two fixed caster wheels, which coincides with the base frame of the bicycle model (positioned at $x(t), y(t)$ in Figure~\ref{fig:nav-kinematics}). The orientation of the coordinate frame in map coordinates is defined by the angle $\Theta$. The distance between this frame and the center of rotation of the diff-drive tractor vehicle is the wheelbase $L$ of the model. The angle of rotation of the tractor vehicle relative to the trailer is the effective steering angle $\delta$.

\paragraph{Global Path Planning}
We use the SBPL (Search-Based Planning Library~\cite{sbpl2023}) Lattice Planner with a rectangular footprint tied to the fixed axis of the trailer. This planner finds the path to the requested goal pose by chaining motion primitives with different lengths and curvatures. By limiting the maximum curvature of the motion primitives the resulting plan is expected to be feasible to be executed by the vehicle via the local planner. The path stubs are generated by a search algorithm and evaluated in an occupancy grid which covers the whole environment.

\begin{figure}[t]
\centering
\includegraphics[height=150pt]{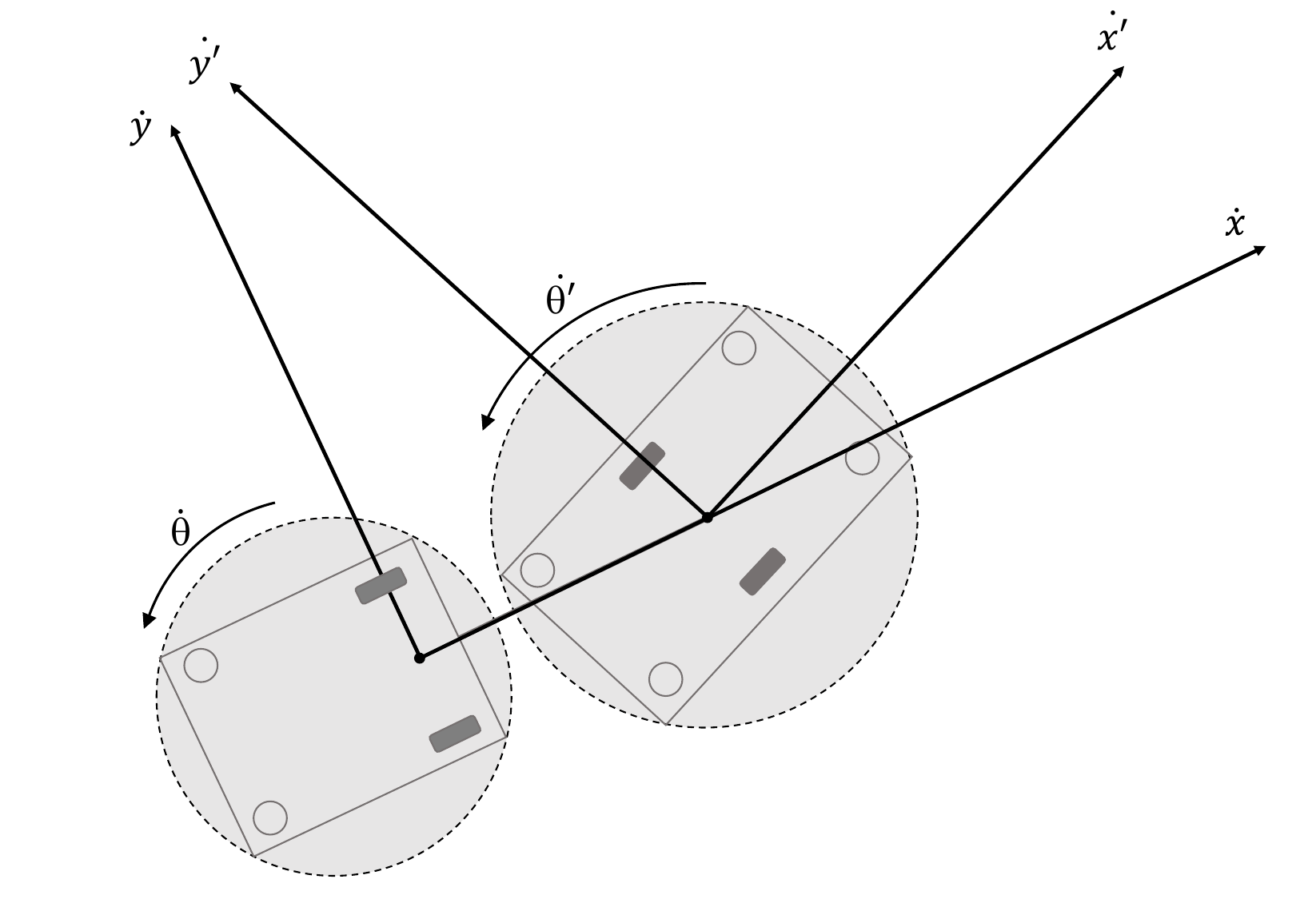}
\caption{\emph{Two Circles} footprint model for the TEB Local Planner (highlighted circles) and the controlled longitudinal and rotational velocities for the car-like vehicle model ($\dot{x}, \dot{\Theta}$) and the tractor vehicle ($\dot{x}', \dot{\Theta}'$). The lateral velocities ($\dot{y}, \dot{y}'$) are zero as they cannot be actuated. }
\label{fig:nav-footprint}
\end{figure}

\paragraph{Local Path Planning}
For the local path following we use the TEB (Time Elastic Band) Local Planner~\cite{Roesmann2012}, which supports navigation for car-like vehicles. For evaluating the actual path, it tracks the obstacles around the vehicle in a smaller, local occupancy grid. It then computes the actual control commands in terms of linear and angular velocities ($\dot{x}$ and $\dot{\Theta}$) to be executed by the vehicle. For representing the system in the local path planner, we use the \emph{Two Circles} footprint model. According to this model, the footprint of the vehicle is specified in terms of two circles that are defined by radius and offset from the vehicle's base frame, respectively (see Figure~\ref{fig:nav-footprint}).

Typically, local planners (e.g., in the ROS navigation stack) provide the output command as the longitudinal speed and angular velocities (here: $\dot{x}, \dot{\Theta}$). To map these commands to the car-like vehicle model we need to add another intermediate controller stage that derives the commands for the tractor vehicle ($\dot{x}', \dot{\Theta}'$), which correspond to the desired behavior of the tractor-trailer system. Given the vehicle's wheel base $L$, the steering angle $\delta$ can be obtained from the target velocities via the curvature of the requested path $\kappa$ with the following formula:

\begin{align}
    \kappa &= \dot{\Theta} / \dot{x} \\
    \delta &= \arctan ( L \kappa )
\end{align}

To control the steering angle, we use a P-controller, which generates the output rotational velocity that would minimize the deviation between the current and the target steering angle:

\begin{align}\label{eq:rotational_vel}
\dot{\Theta}' = K_p \Delta \delta_{normalized}
\end{align}

To avoid issues with angular wrap around, we use the following normalization term to make sure the angle is always in the half-open interval $\left[ -\pi, \pi \right)$: 

\begin{align}
\Delta \delta &= \delta_{target} - \delta_{current} \\
\Delta \delta_{normalized} &= (\Delta \delta + \pi) \bmod 2\pi - \pi
\end{align}

where $\bmod$ is the modulo operation. The current steering angle can be retrieved from the sensor attached to the MiR hook joint. In order to prevent self-collisions, the local planner limits the maximum steering angle the controller will command. Furthermore, we specify the maximum allowed angular velocity to limit the controller output.

In analogy to the angular velocity, the longitudinal velocity is not directly applied to the tractor system. To avoid drift before the desired steering angle is regulated, we add a Gaussian activation function to limit the output longitudinal velocity $\dot{x}'$ based on the current normalized deviation of the steering angle $\Delta \delta_{normalized}$:

\begin{align}\label{eq:longitudinal_vel}
    \dot{x}' = \dot{x} \exp^{-\frac{\Delta \delta^2}{\alpha^2}}
\end{align}

By adjusting the activation factor $\alpha$ we can influence the width of the admissible band of steering angle deviations.
Values of around 10 or larger will practically allow the full longitudinal velocity regardless of the deviation.
Smaller values gradually reduce the bandwidth of the allowed range.

%% file: sections/digital_twin.tex
The K4R platform is an open-source software platform to enable AI and robotics application for retail. At its core, it provides so called semantic digital twins, which are illustrated in the following section. 

\subsection{Semantic Digital Twin}

A semantic Digital Twin (semDT) is a digital representation of a retail store, which connects a \textit{scene graph} to a \textit{semantic knowledge base} as is described in \cite{Kuempel2021}. The \textit{scene graph} contains a 3D model of the store, which is semantically annotated, and holds information like the relative location of the store shelves and products. This data can be automatically generated by a robot driving through the store and scanning all the products within the shelves. The acquired information is connected to an ontology-based \textit{semantic knowledge base}, which is based on interlinked ontologies providing further information on the products, like their ingredients and classifications, 3D models, product taxonomies, product brands or labels. This facilitates semantic reasoning on the semDT, visualization of the 3D environment in various applications, and it allows a human user or a robot to request information using semantic queries. These queries are processed by \textit{KnowRob} \cite{beetz2018know}, which is the underlying knowledge representation and reasoning framework. \\
The Digital Twin itself is constructed using concepts defined in the OWL (Web Ontology Language) format~\cite{OWL}. Internally, everything is represented as a triple, which essentially describes how \textit{entities} are \textit{related} to each other and which \textit{properties} they posses. E.g., \textit{entities}, \textit{relations} and \textit{properties} are the building blocks of the triples. Queries like "which shelves contain empty facings" and "where is a product of type X from the brand Y located" can then be answered in order to help a robot transport products for restocking to the correct locations, or to help guide a customer to a searched product.
\begin{lstlisting}[language=prolog, basicstyle=\scriptsize]
findall(Shelf,
(has_type(Facing, 'http://knowrob.org/kb/shop.owl#ProductFacingStanding'),
\+ triple(Facing, 'http://knowrob.org/kb/shop.owl#productInFacing', _ ),
triple(Facing, 'http://knowrob.org/kb/shop.owl#layerOfFacing', Layer),
triple(Shelf, soma:hasPhysicalComponent, Layer)), Shelves).
\end{lstlisting}
\captionof{lstlisting}{An example query which finds all shelves which contain empty standing facings.}

\begin{lstlisting}[language=prolog, basicstyle=\scriptsize]
subclass_of(ProductType, 'http://knowrob.org/kb/shop.owl#Product'),
has_type(Item, ProductType),
triple(Facing ,'http://knowrob.org/kb/shop.owl#productInFacing', Item),
triple(Facing, 'http://knowrob.org/kb/shop.owl#layerOfFacing', ShelfLayer),
triple(Shelf, soma:hasPhysicalComponent, ShelfLayer).
\end{lstlisting}
\captionof{lstlisting}{A query which returns the location of a product from a certain type (which can be substituted by a brand, for example) in terms of which shelf, layer and facing it is in.}

\begin{figure}
	\centering
	\includegraphics[height=8cm]{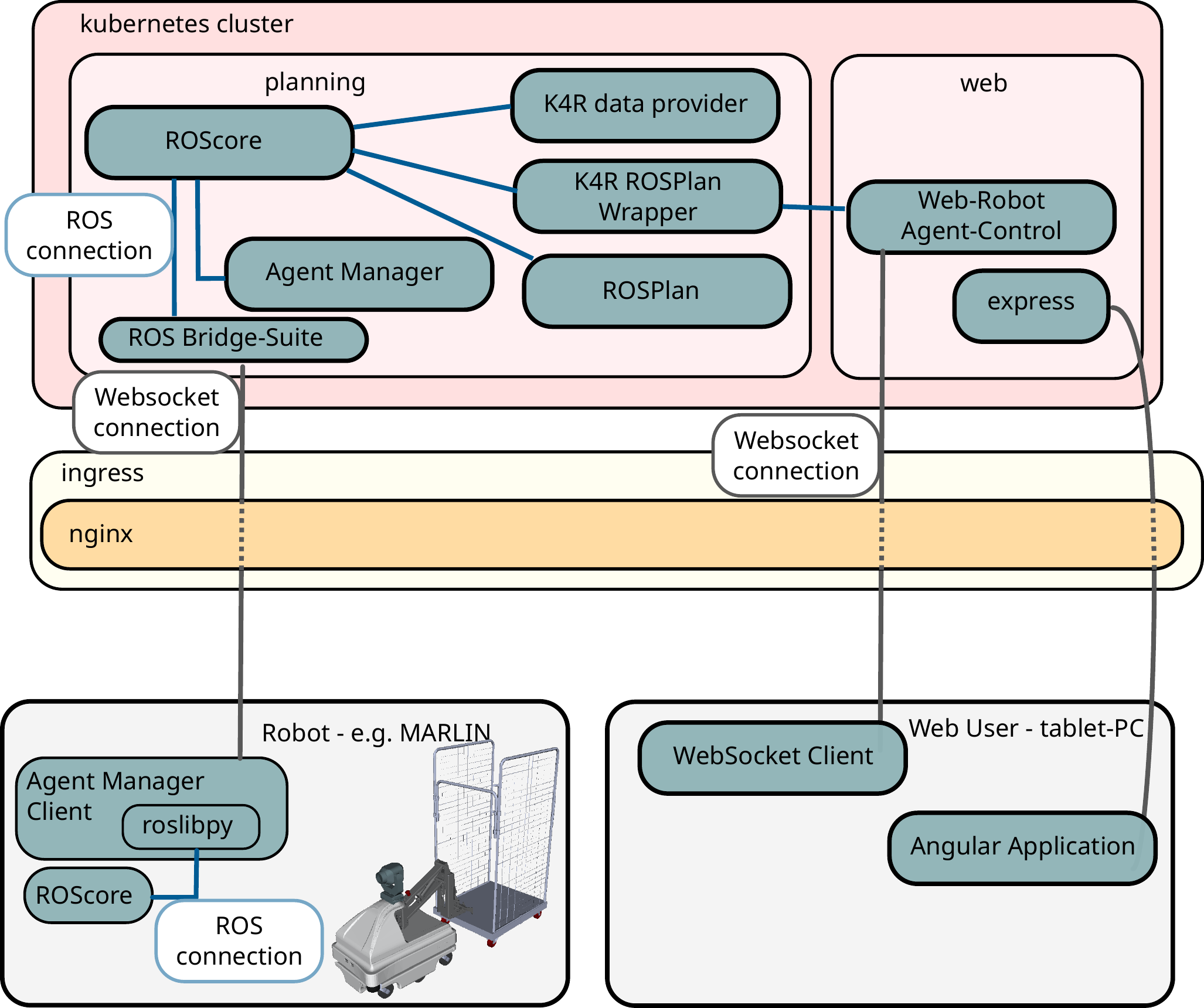}
	\caption{Architecture and interaction of the Web and Planning systems from the users tablet-pc to the robot}
	\label{fig:CloudDeploy}
\end{figure}

\subsection{K4R Platform Architecture and Robot Interfaces}

The K4R infrastructure is developed in containers deployed as pods within a Kubernetes cluster\footnote{\url{https://kubernetes.io/}}. The two main pods described here are called \textit{planning} and \textit{web}. The \textit{planning} pod uses the ROS WebSuite~\cite{rosbridge2023} to connect to robots and other devices via WebSockets. The purpose of the \textit{planning} pod is to use ROSPlan~\cite{Cashmore2015} to create and execute high-level action plans for robotic agents, control the agents, and exchange data with them. The ROSPlan knowledge base is constantly synchronized with the semDT, using the semantic reasoning capabilities of KnowRob. 

The \textit{web} pod is used for human-robot interaction. It hosts a custom Express.js TypeScript application with an Angular front-end through which the user can retrieve information about the agents, launch missions, or directly control one of the robots. The \textit{web} pod sits behind an OAuth2 Keycloak installation to ensure secure communication over the Internet. To allow storing each agent's state, we use certificate-based authentication with tokens and ROSAuth. An overview of the architecture is shown in Figure \ref{fig:CloudDeploy}.
The K4R platform architecture is described in more detail in \cite{K4R}.

\begin{figure}
	\centering
	\includegraphics[height=8cm]{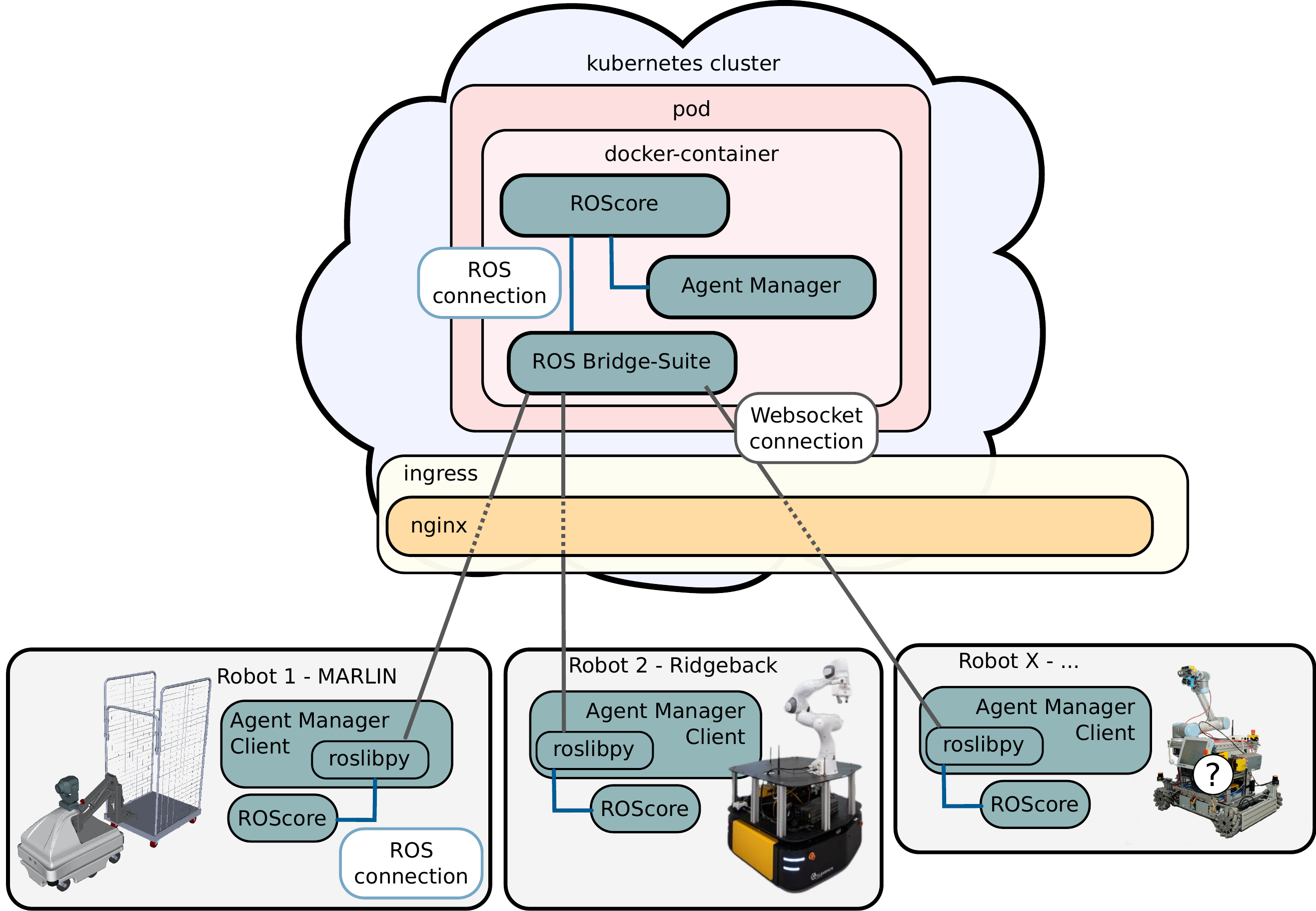}
	\caption{Cloud architecture of Agent Management}
	\label{fig:AgentManager}
\end{figure}

\paragraph{Agent Management} Robots and other agents connect to the K4R platform using the agent manager as shown in Figure~\ref{fig:AgentManager}. It registers with the planning system and provides information about the connected robot. The connection also uses the ROS WebSuite, which is commonly installed on the robotic agent. However, this requires the robot to be fully connected within the network, which is complicated to accomplish when being connected via a cloud platform. Instead of developing a custom server to handle data from the different agents, we deploy the ROS WebSuite on the K4R platform itself, which also brings benefits for robotics software developers, as they can stick to ROS topics and services. 


\paragraph{User Interfaces}
The user interface was developed as a responsive Angular web application that focuses on touch input for tablets, but can also be used with a PC. It allows the user to select a robot, configure it, display information about the system, and start different missions, such as "attach trolley", "go to charging station", or "start refilling the shelves".

Beyond the web user interface, we are developing a smartwatch application for the task of picking. The app notifies the employee when the robot arrives at a delivery point, allowing the employee to place products from the transport cart on the shelves, and confirm when the task is complete.





\paragraph{Task Planning}

For mission planning of \RobotName{} and other robots, we use the ROSPlan planning system from~\citet{cashmore2015rosplan}, which builds on the POPF (Partial Order Planning Forwards) planner from~\citet{coles2010forward}. The planning framework runs on the K4R platform. Thus, it is possible to plan missions with multiple agents and resources, and to easily synchronize the knowledge base of ROSPlan with the semDT. The planning domain is modeled in PDDL (Planning Domain Definition Language~\cite{Ghallab1998}). 

The focus of mission planning here is on the replenishment of the shelves, as well as the related autonomous transport of goods. The goal of such a mission is that all products are autonomously delivered to their target shelves, where they are filled into the shelves by a store employee. In the corresponding domain definition, all robots, trolleys, waypoints for unloading the trolley, as well as the products and their positions must be defined. However, the desired product positions as well as the unloading points can be derived from the store geometry provided by the semDT using KnowRob. The robot autonomously attaches the trolley loaded with products, then moves to the first shelf so that a store employee can replenish it and subsequently confirm the execution on the GUI. This process is repeated until all products are unloaded. The scenario is described as a PDDL domain in Listing \ref{lst:wpnd_pddl}. A problem definition for this domain can be found in Listing \ref{lst:wpnp_pddl}.

The planner does not solve the scheduling and coordination problem, so currently only one agent can be used in a plan. It also ignores the capacity of a cart, so for this scenario we assume that all products can be loaded onto a single cart.


\tiny
\begin{lstlisting}[caption=PDDL domain definition, label=lst:wpnd_pddl]
	(define (domain pick-and-place-domain)

	(:requirements :strips :typing :disjunctive-preconditions :durative-actions)

	(:types
		agent
		waypoint
		product
		trolley
	)

	(:predicates
		; agent reached a waypoint
		(agent_at ?a - agent ?wp - waypoint)
		; defines the order in which the waypoints are visited
		(trolley_at ?t - trolley ?wp - waypoint)
		(docked ?a - agent ?t - trolley)
		(has_trolley ?a - agent)
		(loaded_on ?t - trolley ?p - product)
		(product_at ?p - product ?wp - waypoint )
	)

    (:durative-action dock_trolley
    	:parameters (?a - agent ?t - trolley ?wp - waypoint)
		:duration (= ?duration 10)
    	:condition (and
			(at start (agent_at ?a ?wp))
			(at start (trolley_at ?t ?wp))
		)
    	:effect (and
			(at end (docked ?a ?t))
			(at end (has_trolley ?a))
		)
  	)

    (:durative-action load_on_trolley
    	:parameters (?a - agent ?t - trolley ?wp - waypoint ?p - product)
		:duration (= ?duration 1)
    	:condition (and
			(at start (docked ?a ?t))
			(at start (trolley_at ?t ?wp))
			(at start (product_at ?p ?wp))
		)
    	:effect (and 
			(at end (loaded_on ?t ?p))
			(at end (not (product_at ?p ?wp)))
		)
  	)

	; move to waypoint, without trolley
	(:durative-action move_to_waypoint
		:parameters (?a - agent ?from ?to - waypoint)
		:duration ( = ?duration 10)
		:condition (and
			(at start (agent_at ?a ?from))
		)
		:effect (and
			(at start (not (has_trolley ?a)))
			(at start (not (agent_at ?a ?from)))
			(at end (agent_at ?a ?to)))
	)

	(:durative-action move_to_waypoint_with_trolley
		:parameters (?a - agent ?t - trolley ?from ?to - waypoint)
		:duration (= ?duration 10)
		:condition (and
			(at start (agent_at ?a ?from))
			(at start (trolley_at ?t ?from))
			(at start (docked ?a ?t))
		)
		:effect (and
			(at start (not (agent_at ?a ?from)))
			(at end (agent_at ?a ?to))
			(at start (not (trolley_at ?t ?from)))
			(at end (trolley_at ?t ?to))
		)
	)

	; confirm all products have been placed
	(:durative-action confirm
		:parameters (?t - trolley ?wp - waypoint ?pr - product)
		:duration ( = ?duration 1)
		:condition (and
			(at start (trolley_at ?t ?wp))
			(at start (loaded_on ?t ?pr))
		)
		:effect (at end (product_at ?pr ?wp))
	)
) ; end define
\end{lstlisting}

\begin{lstlisting}[caption=PDDL problem definition, label=lst:wpnp_pddl]
(define (problem pick-and-place-problem)
  (:domain pick-and-place-domain)
  (:objects
    marlin - agent
    trolley0 - trolley
    prod0 - product
    start_area load_area dock_area - waypoint
    wp0 - waypoint
    )
  (:init
    ; in the beginning all agents are in the starting area
    (agent_at marlin start_area)
    ; in the beginning all trolleys are in the docking area
    (trolley_at trolley0 dock_area)
    ; in the beginning all products are in the loading area
    (product_at prod0 load_area)
    )
  (:goal
    (and 
      (product_at prod0 wp0)
	)
  )
)
\end{lstlisting}

\begin{lstlisting}[caption=Basic plan from defined PDDL, label=lst:plan_exec]
0.000: (move_to_waypoint agent0 start_area dock_area)  [10.000]
10.001: (dock_trolley agent0 trolley0 dock_area)  [10.000]
20.002: (move_to_waypoint_with_trolley agent0 trolley0 dock_area load_area)  [10.000]
30.003: (load_on_trolley agent0 trolley0 load_area prod0)  [1.000]
30.004: (move_to_waypoint_with_trolley agent0 trolley0 load_area wp0)  [10.000]
40.005: (confirm trolley0 wp0 prod0)  [1.000]
\end{lstlisting}
\normalsize

%% file: sections/evaluation.tex
In this section, we evaluate the capabilities of \RobotName{} in terms of obstacle detection, autonomous navigation, and task planning. Results are provided in simulation, in laboratory environment, and in a retail store. 

\subsection{Obstacle Detection and Classification}

First, the computational effort of the obstacle detection and classification pipeline is evaluated by measuring the computation time for the different processing steps, as shown in Figure~\ref{fig:sensor_processing_pipeline}. Experimental data is recorded as \RobotName{} navigates a retail store with various obstacles in the aisles (see Figure~\ref{fig:pointcloud_filter_2}). We use two depth cameras with a resolution of 640 x 480 pixels and evaluate the pipeline on \RobotName{}'s onboard PC with 8 x 3.6 GHz, 32 GB RAM. Second, we evaluate the classification performance by measuring the prediction error rate of the classifier using training and test data obtained in a laboratory environment. We use a single depth camera with 640 x 480 pixels in this experiment.

\begin{figure}[t]
  \centering
  \includegraphics[width=\textwidth]{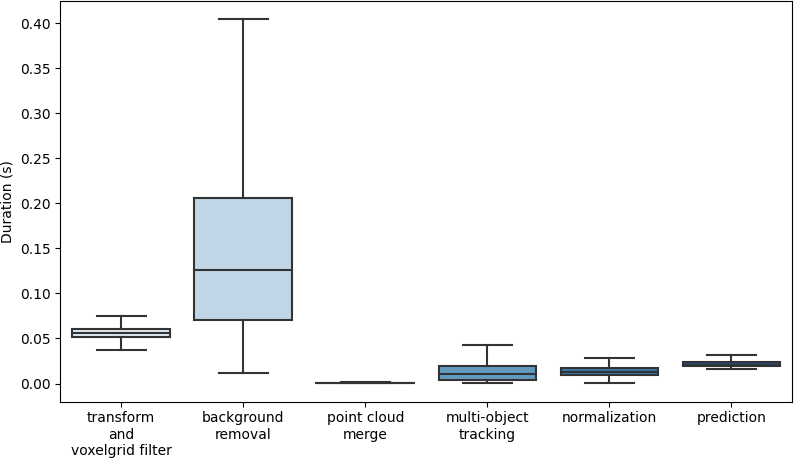}
  \caption{Computation time of the individual processing steps in the obstacle detection and classification pipeline.}
  \label{fig:obstacle_detection_runtime}
\end{figure}


\subsubsection{Computational Efficiency}

To evaluate computational performance of the obstacle detection and classification pipeline, we measure the average computation time of the individual processing steps given $n=1000$ sample points clouds collected when navigating with the \RobotName{} robot in a retail store. The store environment was filled with artificial obstacles like boxes and shopping carts. Figure~\ref{fig:obstacle_detection_runtime} illustrates the results. It can be seen that background removal requires the largest computation time, which is due to the large number of rules that is created for the map of the retail store (see Figure~\ref{fig:map_filter_1}). The poor quality of the map, the diagonal orientation of the shelves, and the way the condition filter works, where rules can only be created parallel to the x- and y-axis, result in the creation of $>1700$ rules in total. The second most time-consuming process is the transformation and voxel grid filter, which depends mainly on the size of the incoming point cloud. The computation times of the other processing steps, namely point cloud merging, tracking, normalization, and preparation, are low in comparison. 

To decrease overall computation time, one could trade off accuracy versus the resolution of the original depth images, manually preprocess the 2D map to produce a lower number of rules in the background filter, or make the clustering step in the multi-object tracking more discriminating to decrease the number of processed clusters.

\subsubsection{Classification Accuracy}

\begin{figure}
  \centering
  \begin{subfigure}[b]{0.19\textwidth}
    \centering
    \includegraphics[width=\textwidth]{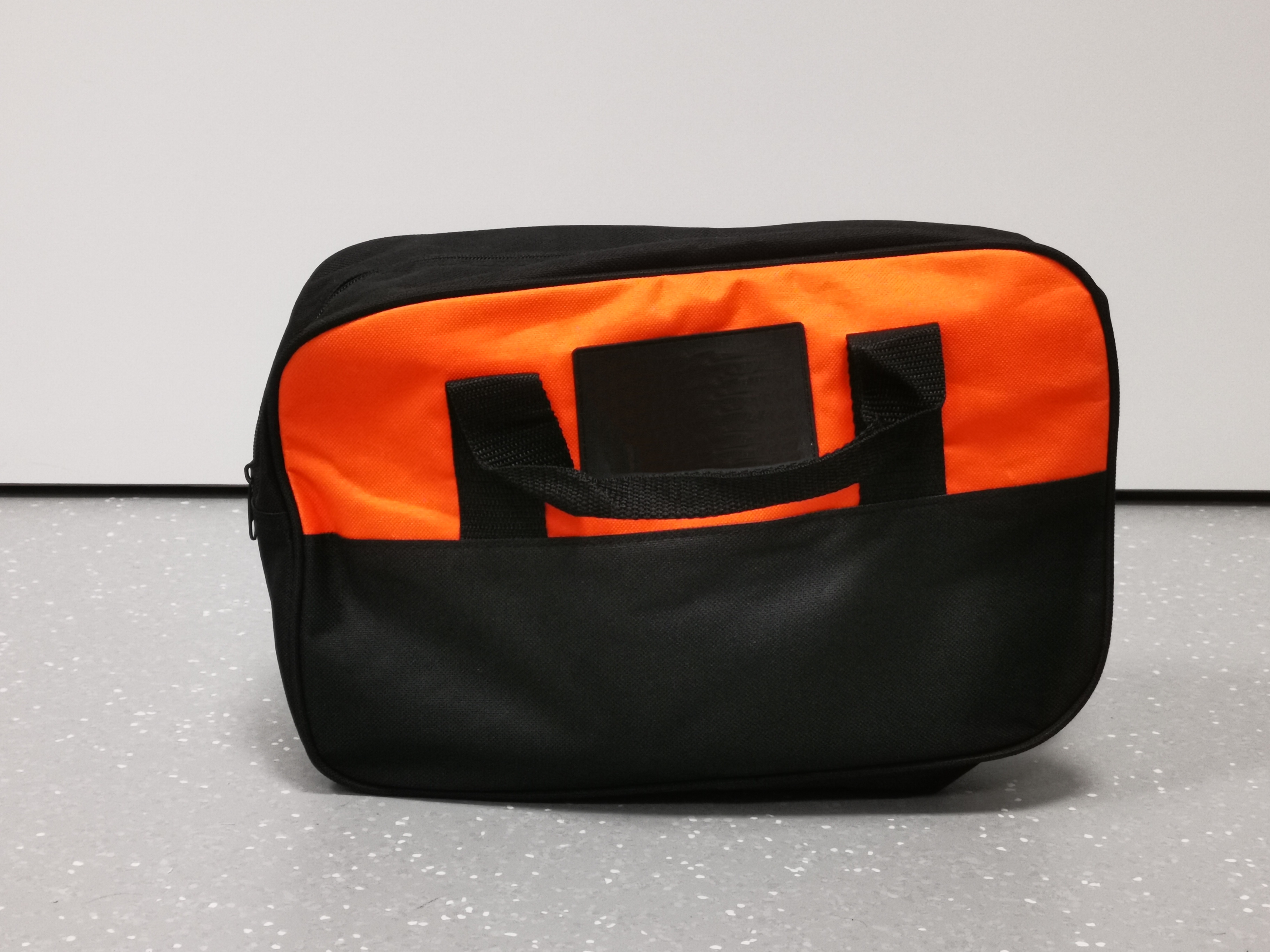}
    \caption{Bag}
    \label{fig:bag_front_view}
  \end{subfigure}
  \begin{subfigure}[b]{0.19\textwidth}
    \centering
    \includegraphics[width=\textwidth]{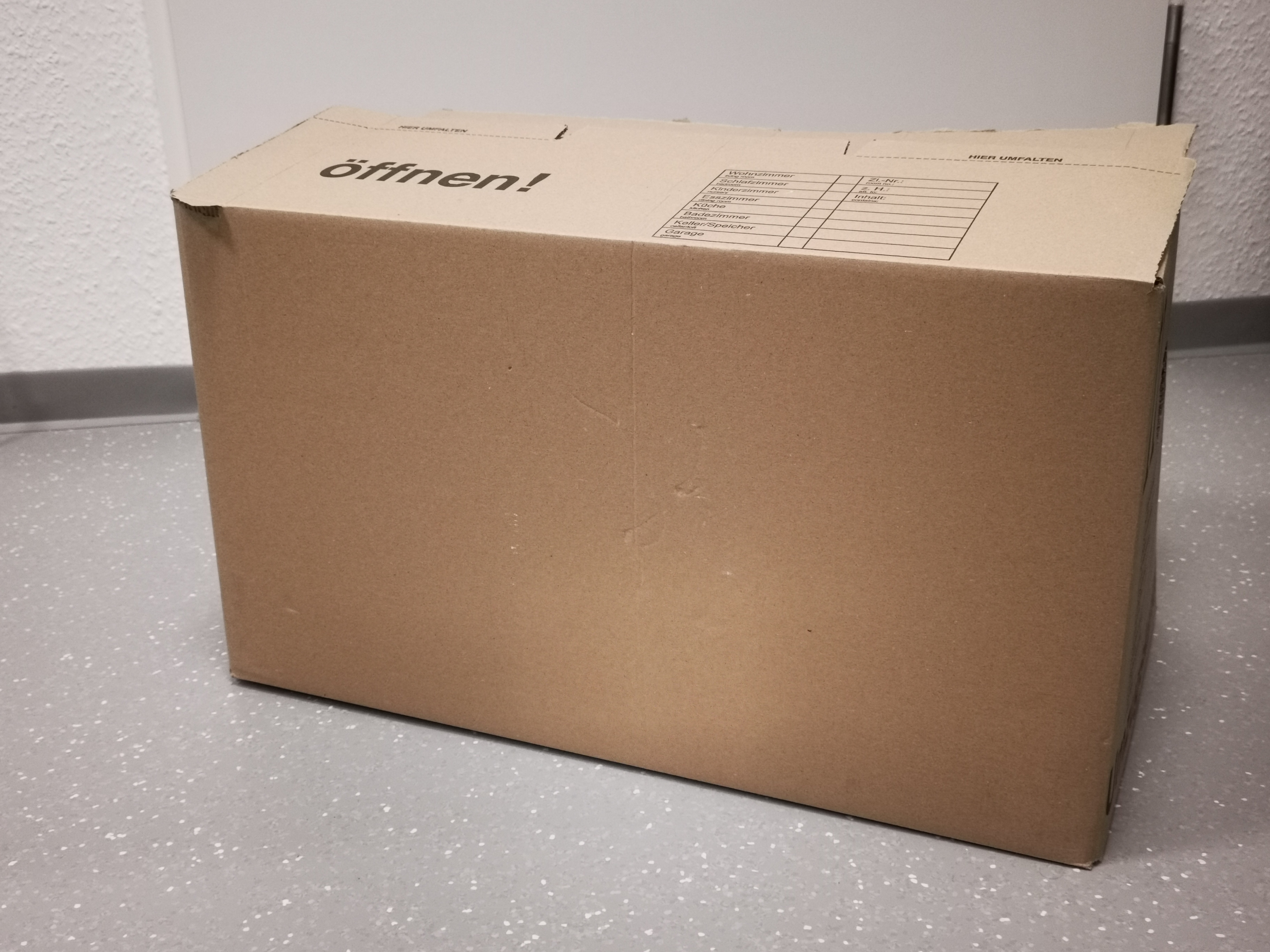}
    \caption{Carton}
    \label{fig:carton}
  \end{subfigure} 
  \begin{subfigure}[b]{0.19\textwidth}
    \centering
    \includegraphics[width=\textwidth]{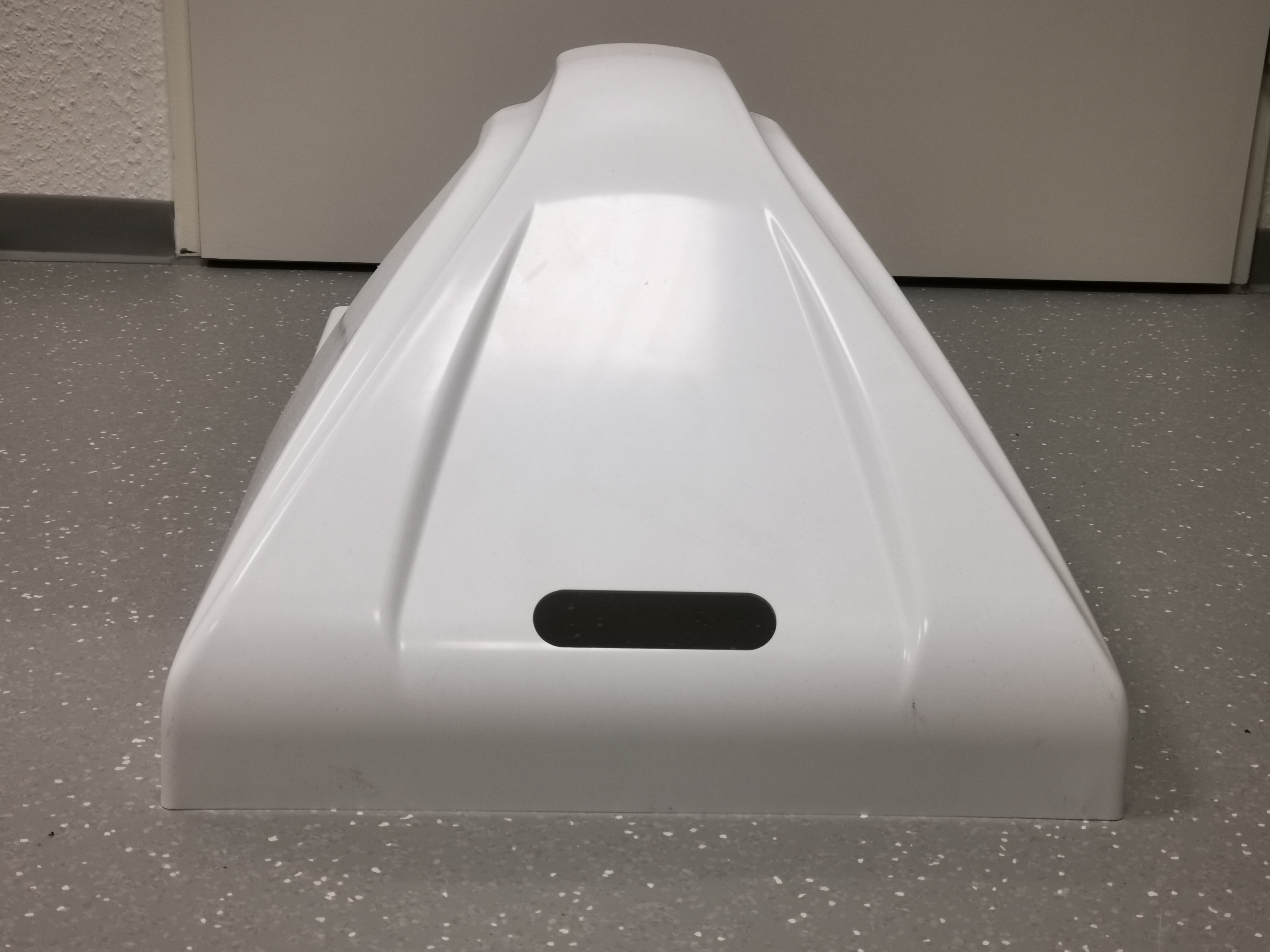}
    \caption{Hook cover front}
    \label{fig:hook_cover_front_view}
  \end{subfigure}   
  \begin{subfigure}[b]{0.19\textwidth}
    \centering
    \includegraphics[width=\textwidth]{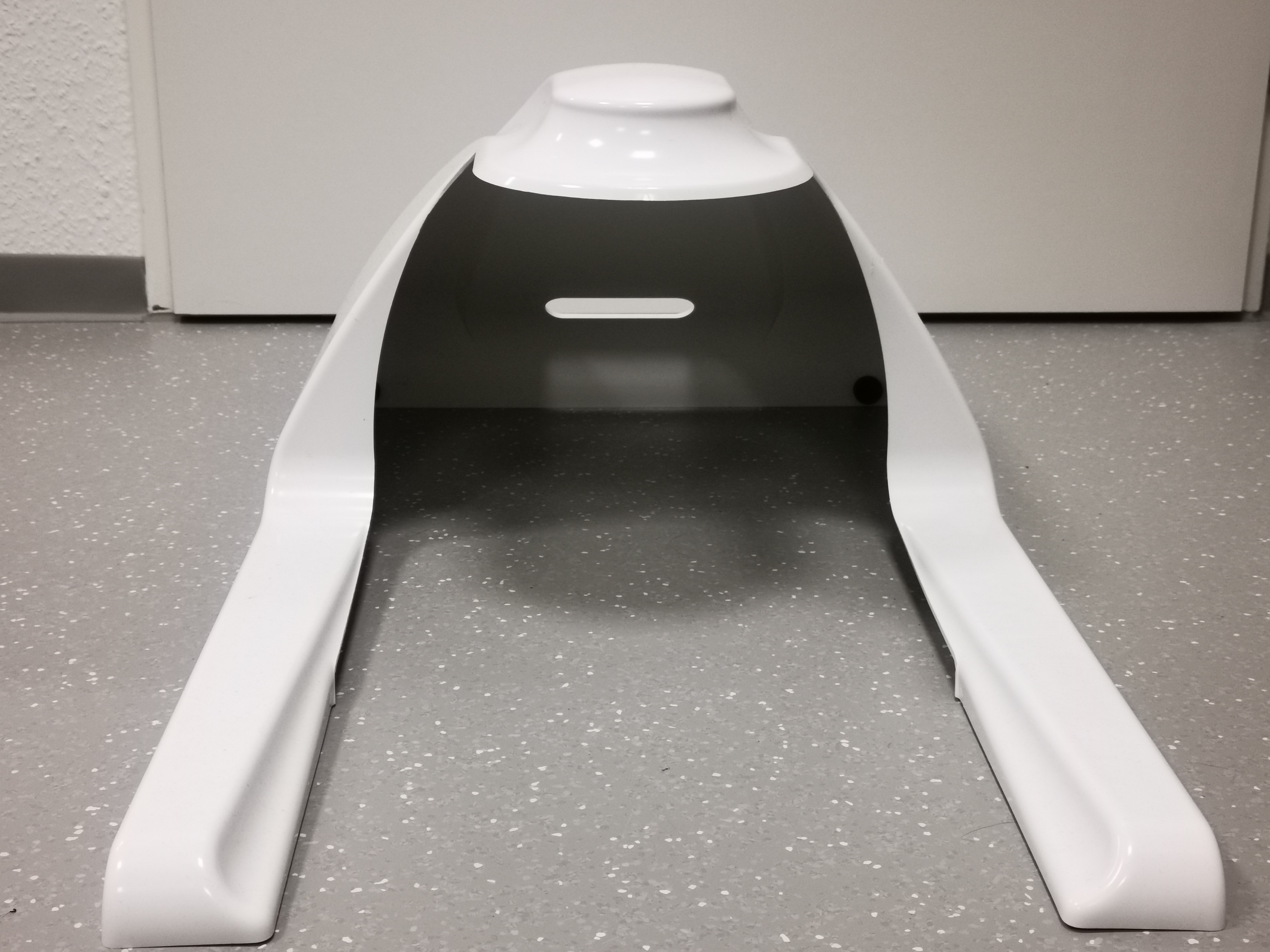}
    \caption{Hook cover rear}
    \label{fig:hook_cover_rear_view}
  \end{subfigure}
  \begin{subfigure}[b]{0.19\textwidth}
    \centering
    \includegraphics[width=\textwidth]{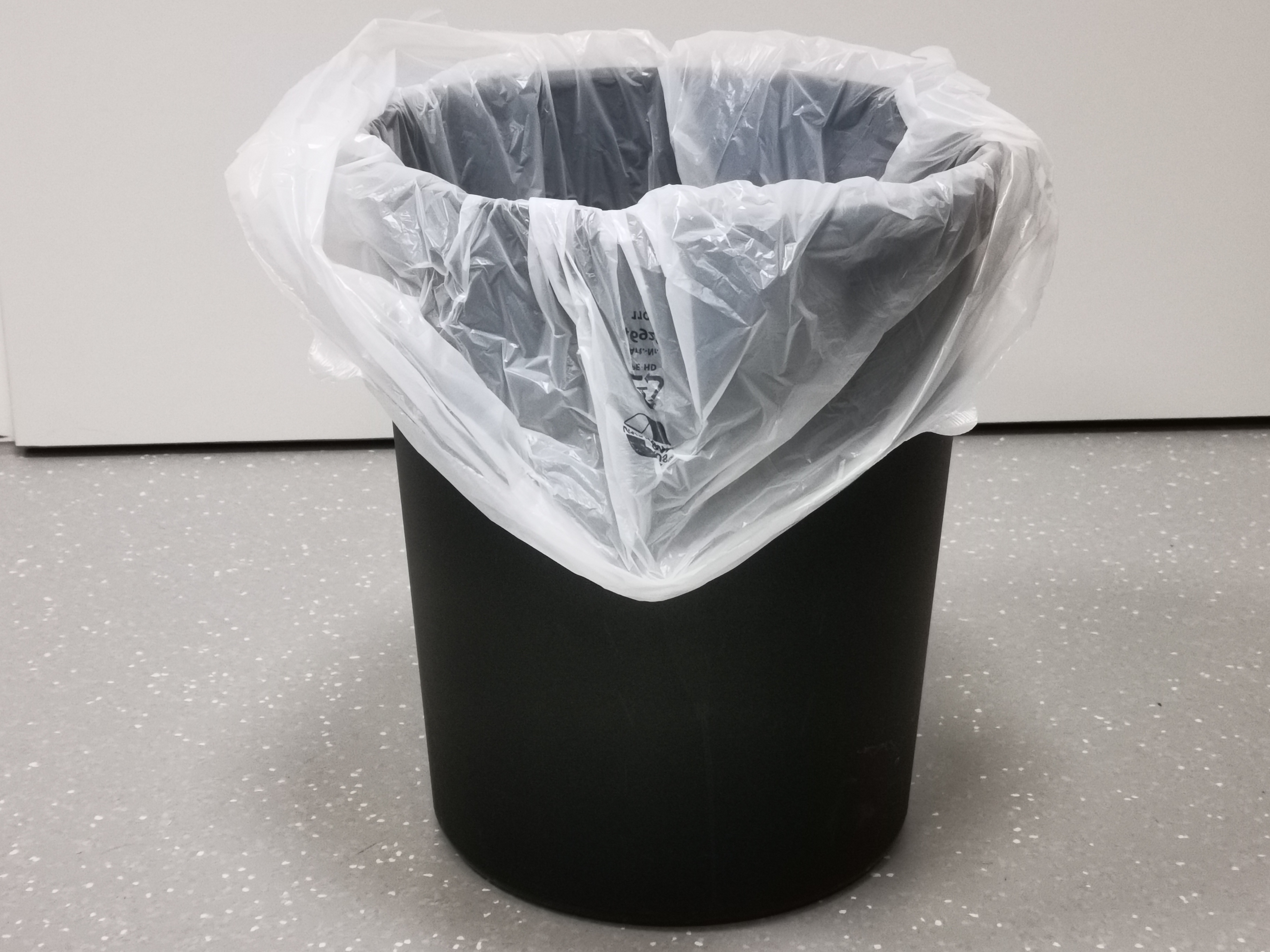}
    \caption{Trash bin}
    \label{fig:trash}
  \end{subfigure}
\caption{Objects used for validation of the obstacle classifier (without human)}
  \label{fig:classification_objects}
\end{figure}

\begin{figure}
  \centering
  \begin{subfigure}[b]{0.49\textwidth}
         \centering
         \includegraphics[height=5cm]{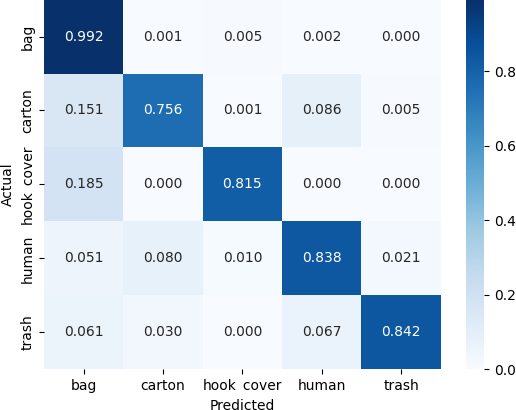}
         \caption{individual predictions}
         \label{fig:conf_matrix_all}
  \end{subfigure}
  \begin{subfigure}[b]{0.49\textwidth}
         \centering
         \includegraphics[height=5cm]{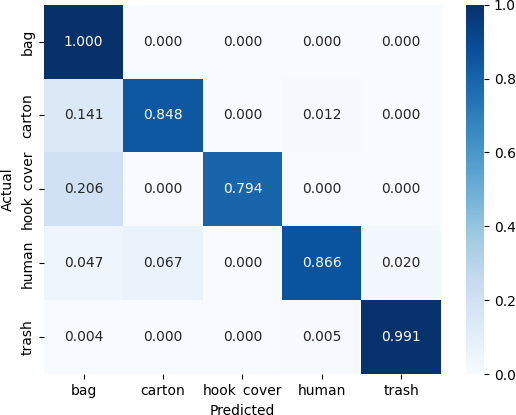}
         \caption{most probable class of a tracked cluster}
         \label{fig:conf_matrix_winner}
  \end{subfigure}
  \caption{Confusions matrices of the SVC (weighted with the probability of the predictions)}
  \label{fig:svc_conv_matrices}
\end{figure}

To evaluate the performance of obstacle classification we record data in a laboratory environment similar to Figure~\ref{fig:sensor_proc_example}. We train 5 different objects (bag, carton, hook cover, human, and thrash can) using raw point cloud data. To evaluate the error rate of the trained SVC model, we consider both the individual predictions of the model (Figure~\ref{fig:conf_matrix_all}) and the output prediction (Figure~\ref{fig:conf_matrix_winner}), where ten predictions are combined into one. The validation of the trained model was performed with live data from objects of all classes shown in Figure~\ref{fig:classification_objects}. Except for the classification of the hook cover (which is sometimes split into two clusters due to its shape, leading to false classifications), the precision of the classification can be improved by considering ten classifications.


\subsection{Tractor/Trailer Navigation}


The capabilities of the approach for tractor-trailer navigation are first evaluated in simulation. In the next step, we reproduce the results on the real system in a similar environment and compare our approach with the capabilities of the proprietary navigation stack.

\subsubsection{Evaluation in Simulation}
\begin{figure}[ht]
  \centering
  \includegraphics[height=240pt]{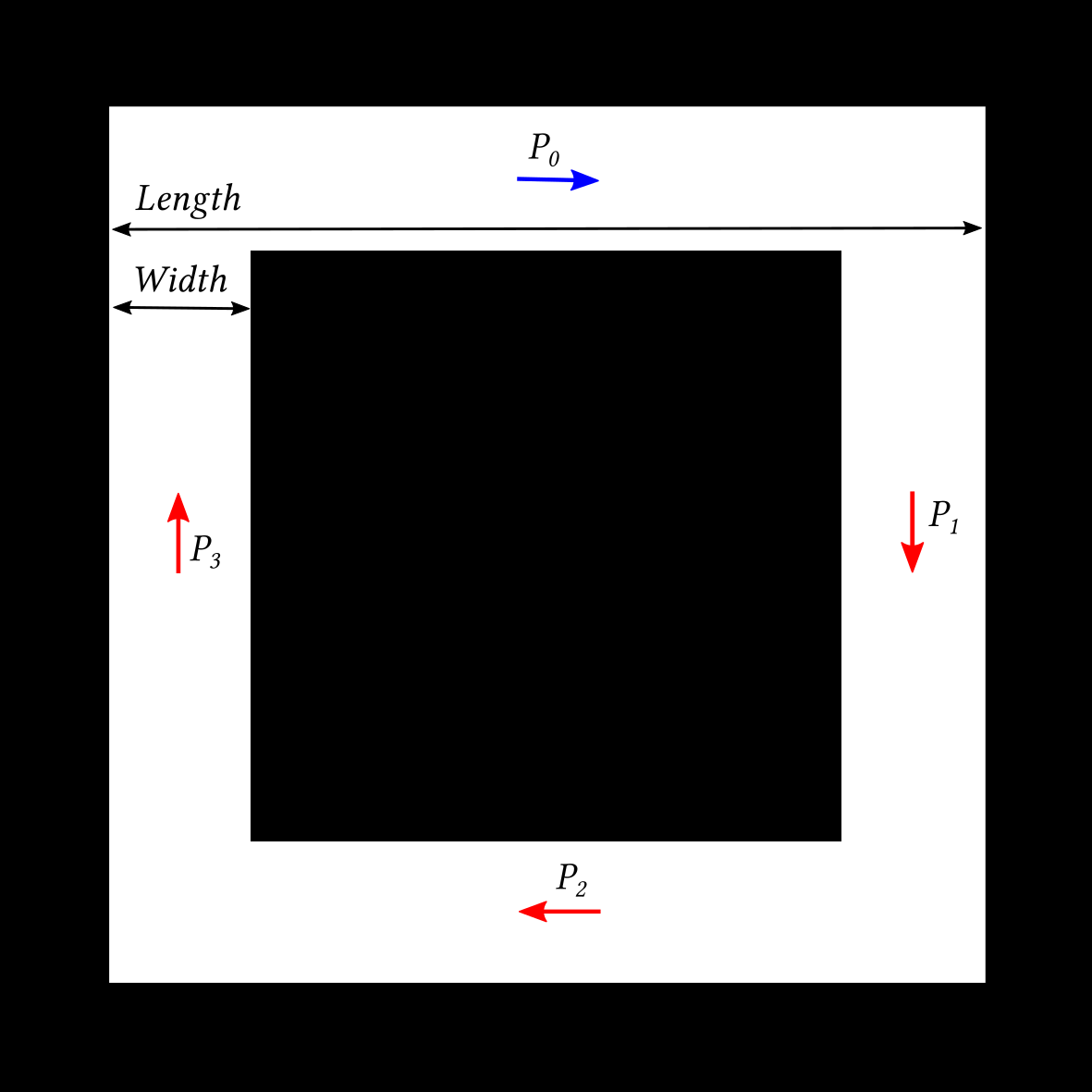}
  \caption{Schematic top view of the artificial simulation environment for evaluation of the navigation capabilities.}
  \label{fig:exp_env1}
\end{figure}

We first create a simple simulation environment in which the robot is to navigate along a corridor. Figure~\ref{fig:exp_env1} shows a schematic top view of the environment, with the walls in black and the empty floor space in white. 
The central square wall is resized in steps of \SI{0.1}{\meter} to create different corridor widths between \SI{1.4}{\meter} and \SI{2.0}{\meter}. 
The values were selected based on the corridor widths typically found in retail stores.
The corridor length is kept constant at \SI{10}{\meter}. 
The target points $P_0$ - $P_3$ are located in the middle of the respective sides and are aligned so that the robot only has to move forward.

\paragraph{Procedure}

Initially, the robot is placed at position $P_0$ and is then asked to navigate to the positions $P_1$, $P_2$, $P_3$, and $P_0$ in order. 
The goal tolerance of the navigation approach is selected to allow \SI{0.5}{\meter} of translational and \SI{0.2}{\radian} of rotational deviation. 
We run the experiment 25 times for each configuration.
If the robot fails to reach a position, we register this failure. 
In this case, the next position is chosen nevertheless, so that all goal positions are evaluated on each run. 
We justify this procedure as follows: Sometimes, despite an error at one position, the robot still manages to reach the subsequent positions. 
This happens either because the robot drives past an unreachable position, or because it drives backwards from a corner where it was stuck before.
The entire run is aborted if the laser scanner detects an obstacle in the circular safety zone around the towing vehicle that is slightly larger than its actual footprint.
We record the trajectories of the tractor $x'(t), y'(t)$ and the trailer coordinate system $x(t), y(t)$ during the experiments, as well as the execution time, and whether each position has been reached or not.
The goal of evaluation is to measure the success rate (in terms of the number of positions reached successfully) and the average duration for each run as a function of corridor width to get an idea of the expected performance on the real system.

\paragraph{Results}

\begin{figure}[htb]
  \centering
  \includegraphics[height=240pt]{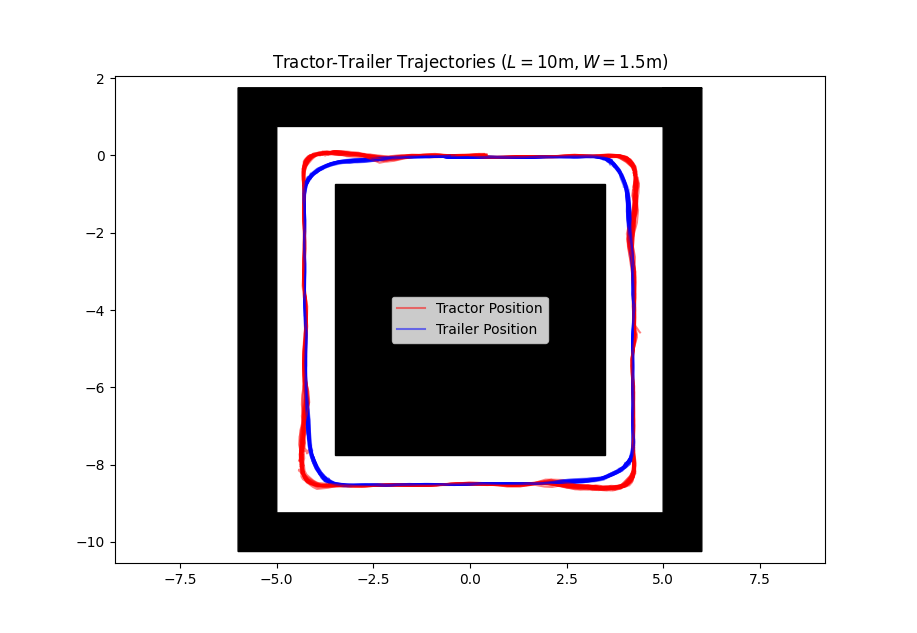}
  \caption{Visualization of tractor and trailer trajectories.}
  \label{fig:results:tractor-trailer_trajectories}
\end{figure}

Figure \ref{fig:results:tractor-trailer_trajectories} shows the trajectories of the tractor and trailer positions along the track. It can be seen that
the tractor overshoots the center path at the corners, because otherwise the trailer would collide with the inner walls.

\begin{figure}[htb]
  \centering
  \begin{subfigure}[c]{0.49\linewidth}
\centering
  \includegraphics[height=4.1cm]{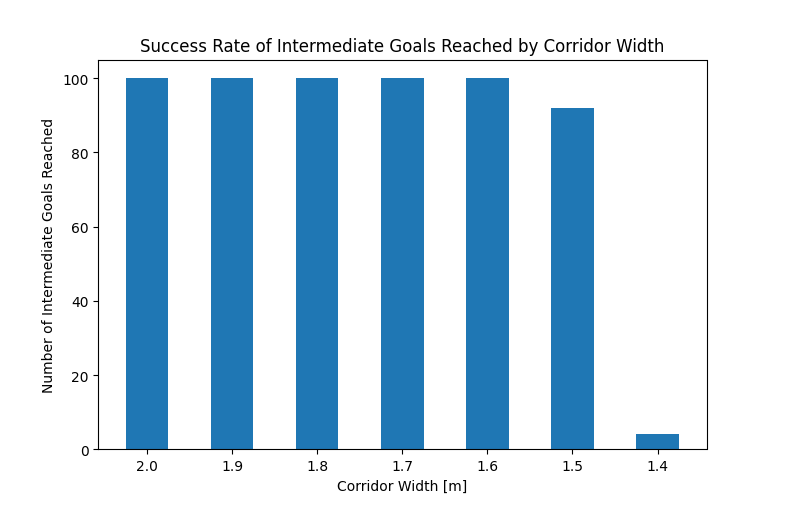}
  \caption{Success rate in reaching the target positions.}
  \label{fig:results:success_rate}
  \end{subfigure}
\hfill
  \begin{subfigure}[c]{0.49\linewidth}
\centering
  \includegraphics[height=4.1cm]{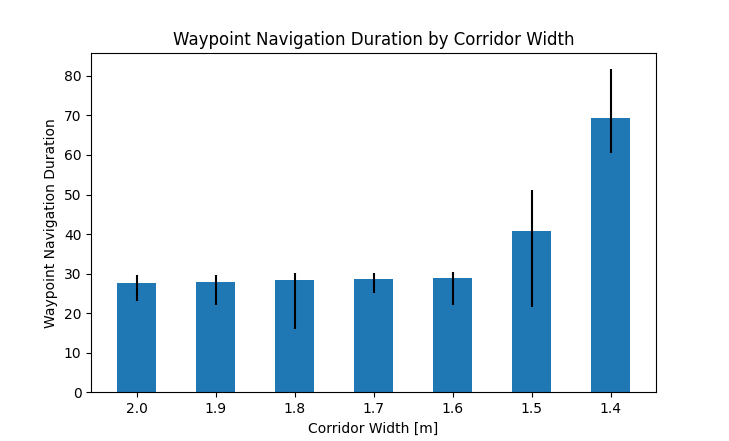}
  \caption{Navigation time per target}
  \label{fig:results:navigation_duration}
  \end{subfigure}
    \caption{Results on tractor-trailer navigation in simulation.}
\end{figure}

Figure~\ref{fig:results:success_rate} illustrates the number of intermediate targets reached along the route over the corridor widths. It can be seen that the approach works reliably down to a corridor width of \SI{1.6}{\meter}.
With a corridor width of \SI{1.5}{\meter}, 92 out of 100 intermediate targets are still reached during the 25 passes.
At a corridor width of \SI{1.4}{\meter}, the success rate drops sharply. Here, the vehicle reaches the first intermediate target in only four of 25 passes.

In addition, we evaluate the time required to navigate around a single corner, which naturally increases as corridor width decreases.
This is evident for corridor widths of 1.5 meters and less, while the time required to navigate the route is relatively constant for corridor widths of 1.6 meters and more (see Figure~\ref{fig:results:navigation_duration}).

\subsubsection{Real-World Evaluation}
\begin{figure}[ht]
\centering
\begin{subfigure}[c]{0.68\linewidth}
\centering
  \includegraphics[height=6.5cm]{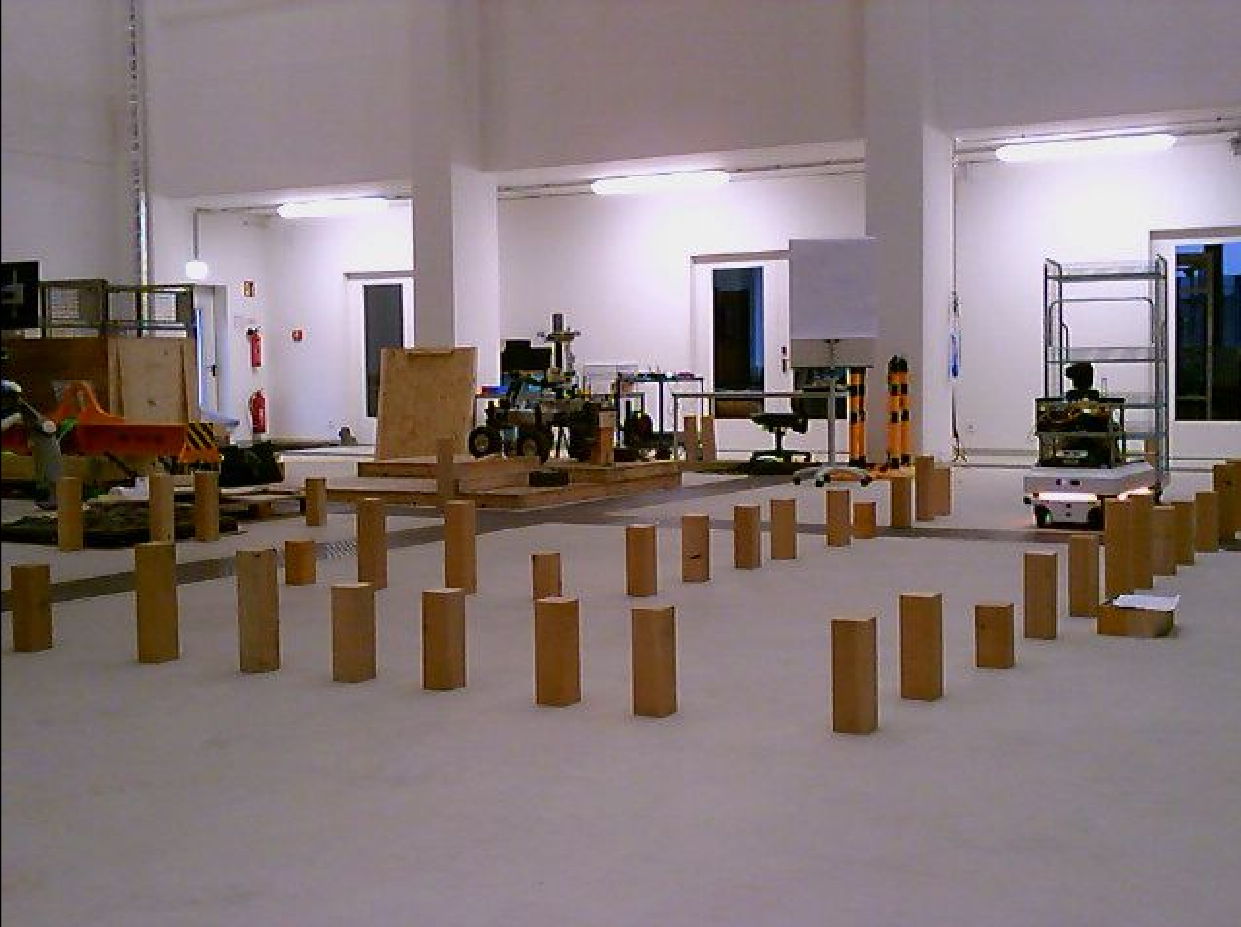}
  \caption{Perspective view of the actual setup. The walls are made up of wooden blocks that can be moved to adjust the corridor width.}
 \label{fig:exp_env2:picture}
 \end{subfigure}
\hfill
\begin{subfigure}[c]{0.28\linewidth}
\centering
\includegraphics[height=6.5cm]{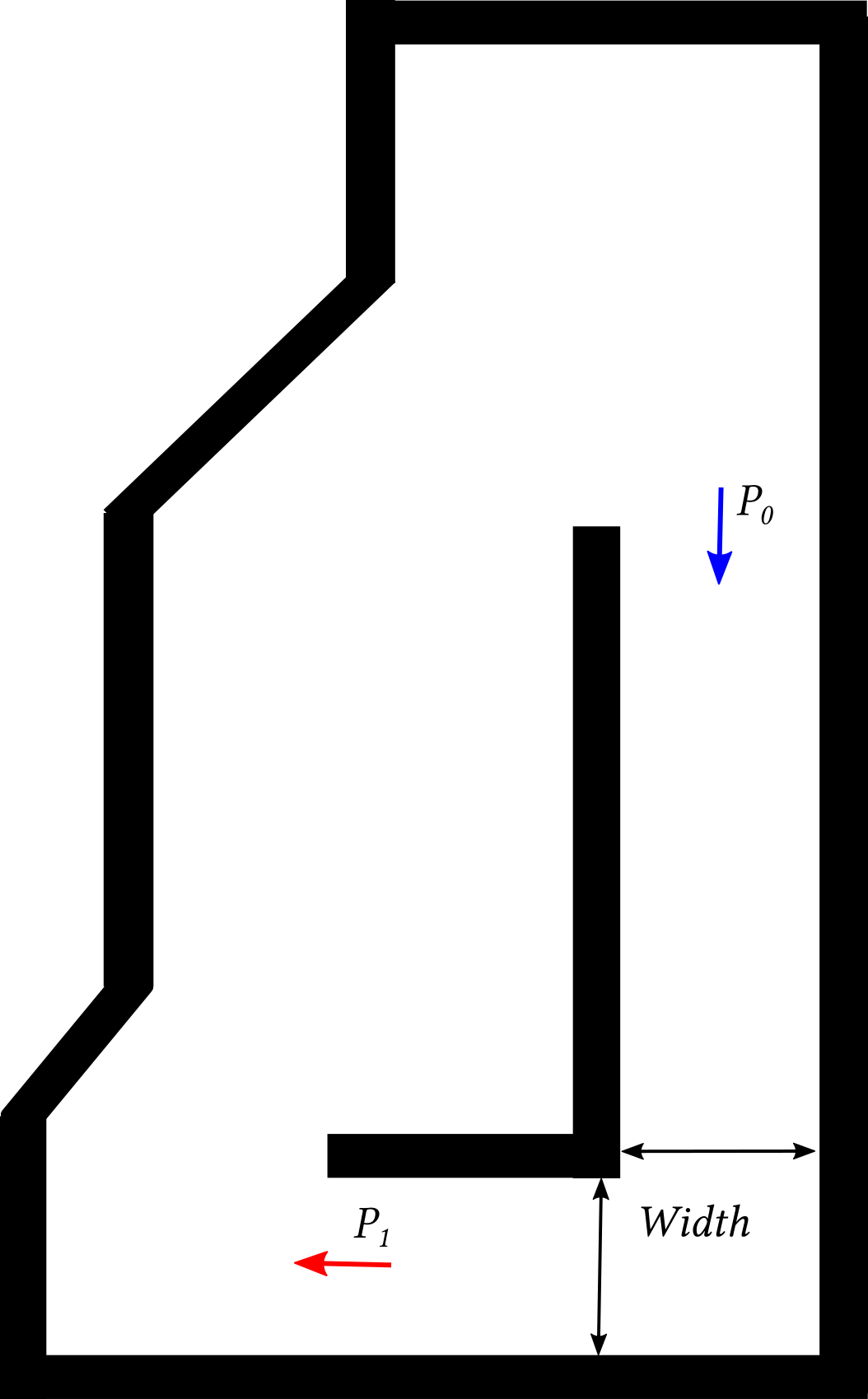}
\caption{Schematic top view of the evaluation environment.}
\label{fig:exp_env2:schematic}
\end{subfigure}
\caption{Real-World Evaluation.}
\label{fig:exp_env2}
\end{figure}

For the evaluation on the real system, we create a similar track as in simulation.
However, unlike the simulation, the corridor has only one corner with critical width, while the rest of the path is wide enough to easily return to the starting position before the next pass.
Figure~\ref{fig:exp_env2:schematic} shows a schematic overview of the setup with the walls in black and the free area in white. The central wall is moved to create different corridor widths. The positions $P_0$ and $P_1$ are adjusted accordingly to always be in the center of the corridor.
The goal of this evaluation is to reproduce the performance observed in the simulation and compare our approach with the proprietary navigation method of the MIR robot, which is provided by the manufacturer.

\paragraph{Procedure}

We start with a corridor width of \SI{1.9}{\meter} and reduced it in increments of \SI{0.1}{\meter}.
The goal tolerance was set to allow \SI{0.5}{\meter} translation and \SI{0.2}{\radian} rotation error.
For each corridor width, we start the experiment with the robot in pose $P_1$ and send it to the target poses $P_0$ and $P_1$ alternately, regardless of whether the previous target was reached or an error occurs.
We repeat the evaluation five times without manual intervention for both our custom tractor-trailer navigation approach and the proprietary navigation stack of the MIR robot.
The run is aborted if the system hits an obstacle and is halted by its own safety stop system or if the emergency stop has to be pressed by the human operator.
We record the timestamps at which the target positions are reached or, in case of a target position is not reached, we register an error.


\paragraph{Results}

Table \ref{tab:exp:realworld} shows the success rate of the approaches with respect different corridor widths.
\begin{table}[htb]
\centering
\begin{tabular}{c|cc}
Corridor Width & Custom & Proprietary \\
\hline
\SI{1.9}{\meter} & 4 / 5 & 5 / 5 \\
\SI{1.8}{\meter} & 5 / 5 & 5 / 5 \\
\SI{1.7}{\meter} & 5 / 5 & 5 / 5 \\
\SI{1.6}{\meter} & 4 / 5 & 5 / 5 \\
\SI{1.5}{\meter} & 5 / 5 & 0 / 5
\end{tabular}
\caption{Success rate for reaching the goal pose $P_1$ in real-world evaluation for custom and proprietary navigation approach.}
\label{tab:exp:realworld}
\end{table}

We find that the proprietary navigation approach is able to reliably navigate the course down to a corridor width of \SI{1.6}{\meter}.
However, at a corridor width of \SI{1.5}{\meter}, the planner is no longer able to find a path through the course.
The custom navigation approach, on the other hand, is able to navigate the course up to and including a width of \SI{1.5}{\meter}.

Also, in the custom navigation approach, we observe two error cases at corridor widths of \SI{1.9}{\meter} and \SI{1.6}{\meter}. Both occur at the turn just before the start position $P_1$, where the vehicle gets stuck and aborts the run. 
It then continues with the subsequent targets, which are again successfully reached.

\subsubsection{Discussion}

Using the proposed method for tractor-trailer navigation, the robot can navigate reliably up to a corridor width of \SI{1.6}{\meter} in the simulated environment.
Below \SI{1.5}{\meter}, navigation time increases noticeably and the system begins to occasionally get stuck at a corner.
At even lower corridor widths, the success rate drops to near zero, which is consistent with the manufacturer's stated limitations for the system.

In the real-world robot application, the custom approach to tractor-trailer navigation was able to navigate narrower corridors than the MIR robot's proprietary approach. 
However, the proprietary navigation approach tends to deliver smoother and more robust execution, possibly due to better fine-tuning of navigation parameters. 
With manual control of the robot, even much smaller corridor widths are possible, leaving room for further optimization for the autonomous navigation.

\subsection{Task Planning}

\begin{figure}
	\centering
	\includegraphics[height=7cm]{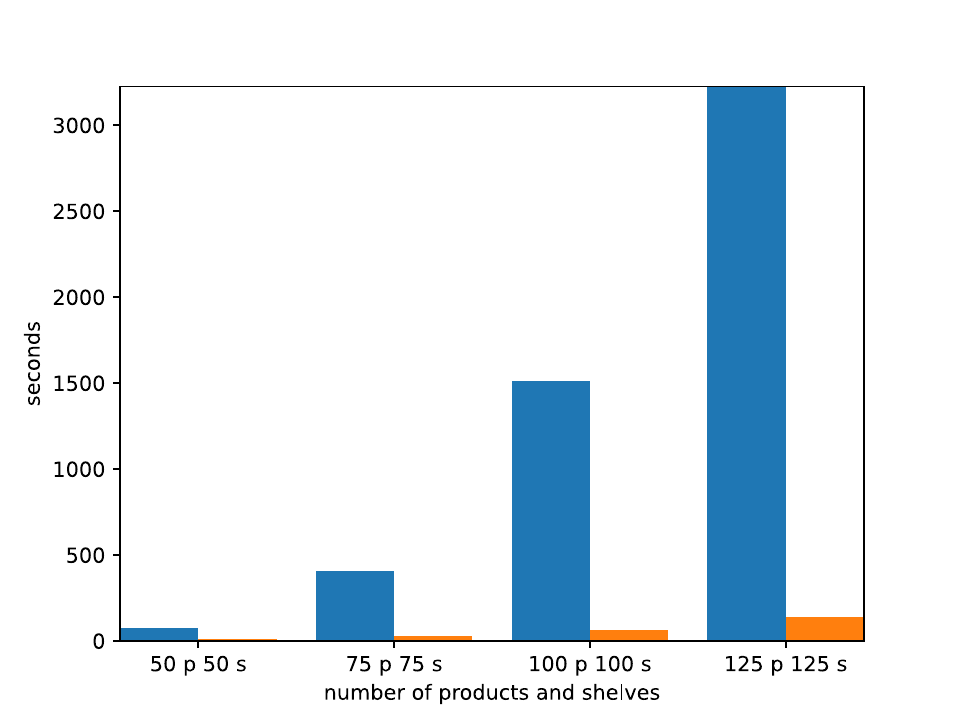}
	\caption{Planning time based on amount of products (p) and shelves (s). Blue bars - No prior knowledge on the order of the unloading points, orange bars - Order between the unloading points is pre-defined.}
	\label{fig:PlanningTimeProductsShelves}
\end{figure}

In this section, we evaluate the performance of the planner with respect to the size of the store and the number of products. For this purpose, we use the domain description shown in Listing \ref{lst:wpnd_pddl} and the problem definition shown in Listing \ref{lst:wpnp_pddl}. This definition describes the task "Replenish all items loaded on the cart" in PDDL. We assume here that the robot can approach all available unloading points without exceptions. All computation is performed on an Intel i7-8550U CPU.
We use the POPF planner from ROSPlan and evaluate different transportation scenarios. As can be seen in Figure \ref{fig:PlanningTimeProductsShelves}, the planning time increases exponentially with the number of products and shelves (blue bars). Thus, a completely free definition where the agent can move to any waypoint is not practical in a realistic scenario with a few hundred products and over a hundred shelves, which requires narrowing the search space.

If we give the planner a graph with a fixed order in which the waypoints must be approached, the planning time is reduced considerably, as can be seen in Figure \ref{fig:PlanningTimeProductsShelves} (orange bars). Thus, we use a simple heuristic from a distance matrix of the store to get an initial guess of the unloading order. The distance matrix is fixed for each store and can be obtained using the store layout. It contains the pairwise distances between all relevant locations in the store, e.g., the shelves.

If an unloading point cannot be approached (e.g., because an obstacle is in the way), and the navigation planner of the robot cannot find an alternative route, the sequence of unloading points is adjusted.  In this case, the robot initially omits the next unloading point, adds it to the end of the list as an additional destination, and proceeds to the next unloading point. It is assumed that the unloading point is only temporarily blocked (e.g. due to high customer traffic in an aisle) and will be available again after some time.

It will be very beneficial to provide additional information about the specific store in the knowledge base of the planner. If the obstacle can not move the robot might need to call a shop employee for help but if there is a human in the path it could ask the human politely to move or ask a shop employee for help. An option to improve the robots' behavior would be to use Reinforcement Learning to figure out where these areas are and optimize the order in which the waypoints are approached accordingly. A possible implementation for this has been investigated by \cite{liu2022}, but needs to be validated outside of simulation and is not integrated into the K4R platform.

\subsection{Use Case: Support of Shelf Refilling}
\label{subsec: Replenishment}

\begin{figure}
\centering
\begin{subfigure}[t]{\textwidth}
\includegraphics[width=0.32\columnwidth]{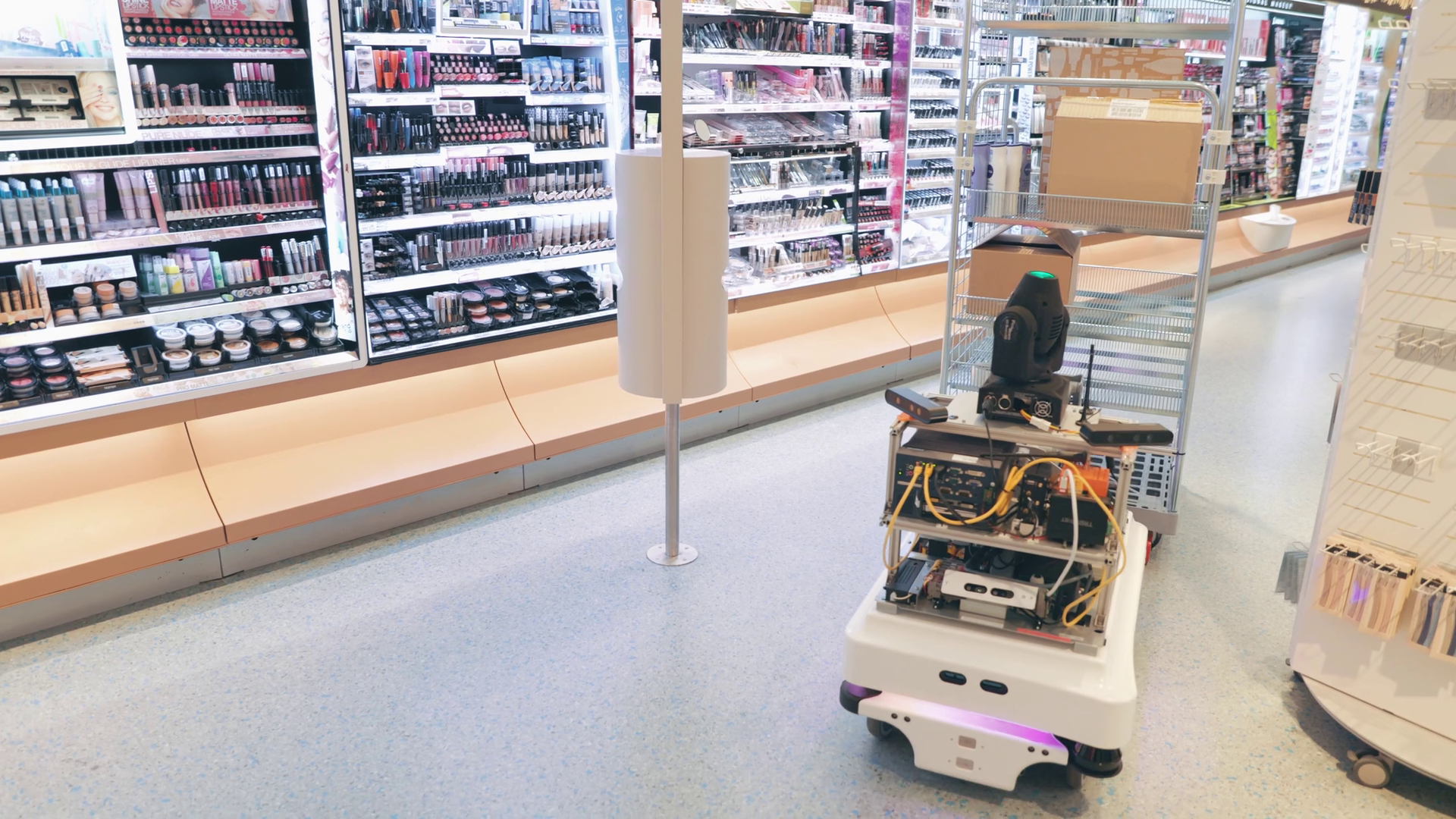}\hfill
\includegraphics[width=0.32\columnwidth]{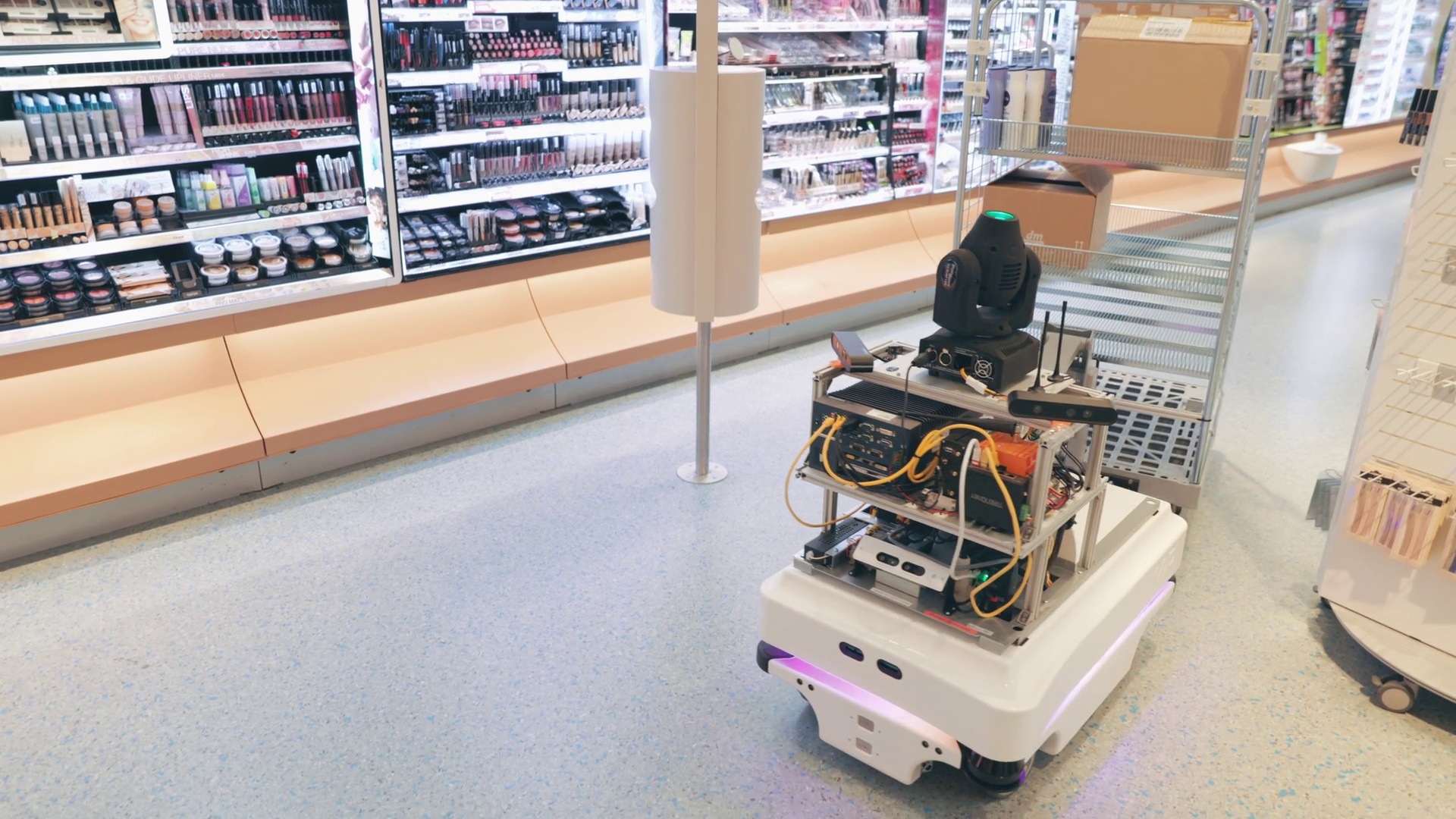}\hfill
\includegraphics[width=0.32\columnwidth]{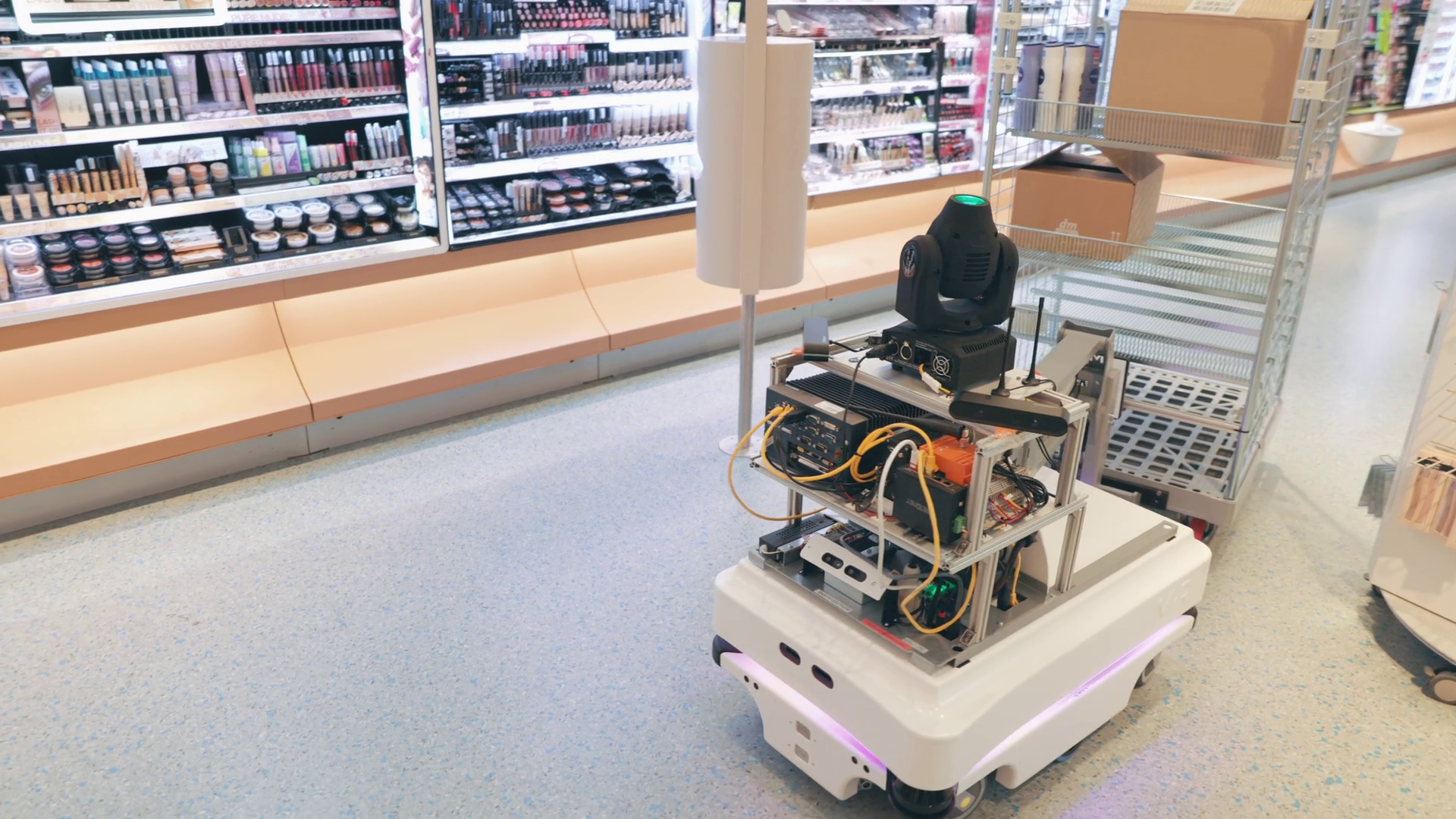}
\caption{\RobotName{} navigating narrow passages in a retail store.}
\label{fig:use_case_1}
\end{subfigure} \vspace{0.3cm} \\

\begin{subfigure}[t]{\textwidth}
\includegraphics[width=0.32\columnwidth]{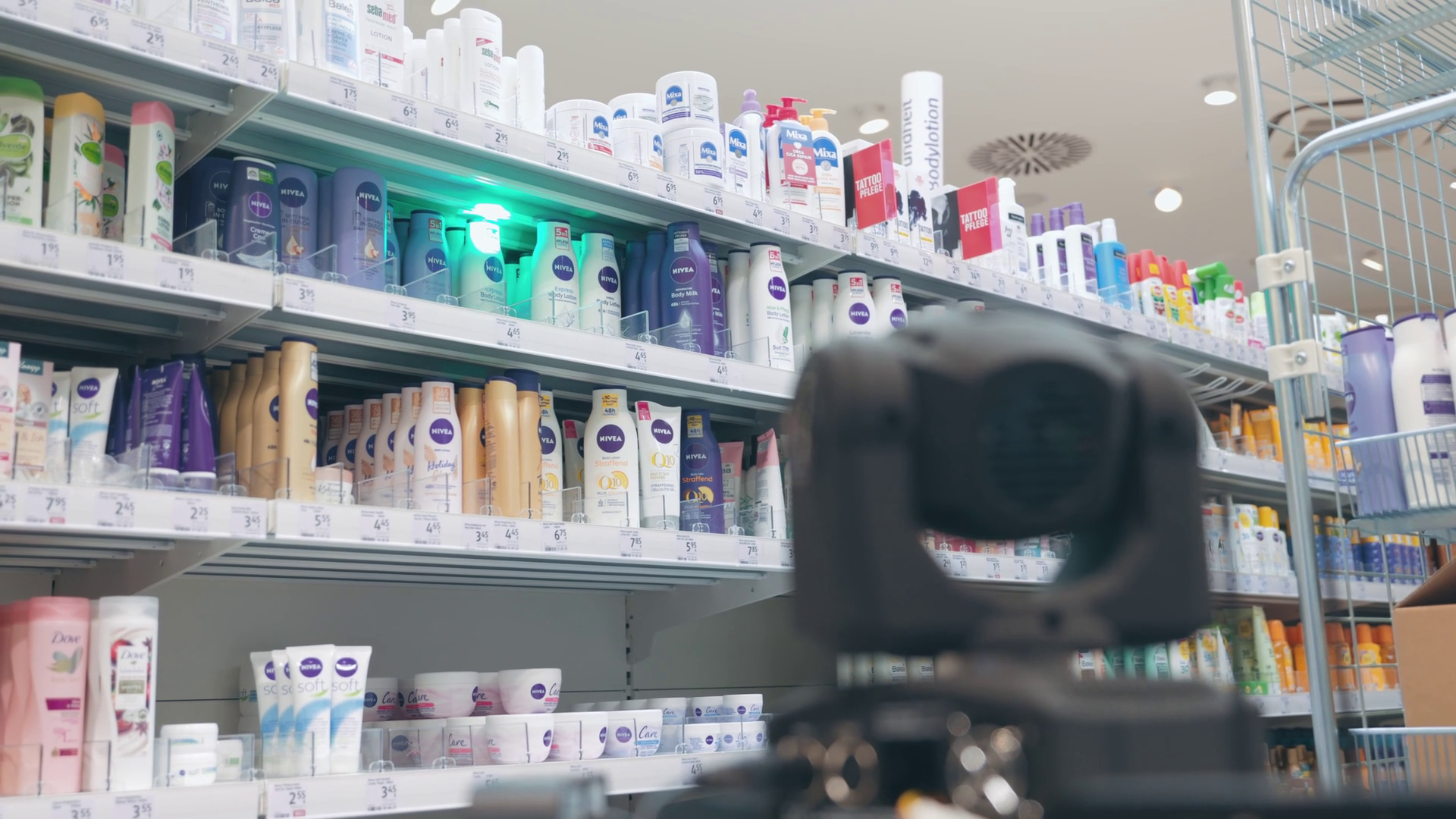}\hfill
\includegraphics[width=0.32\columnwidth]{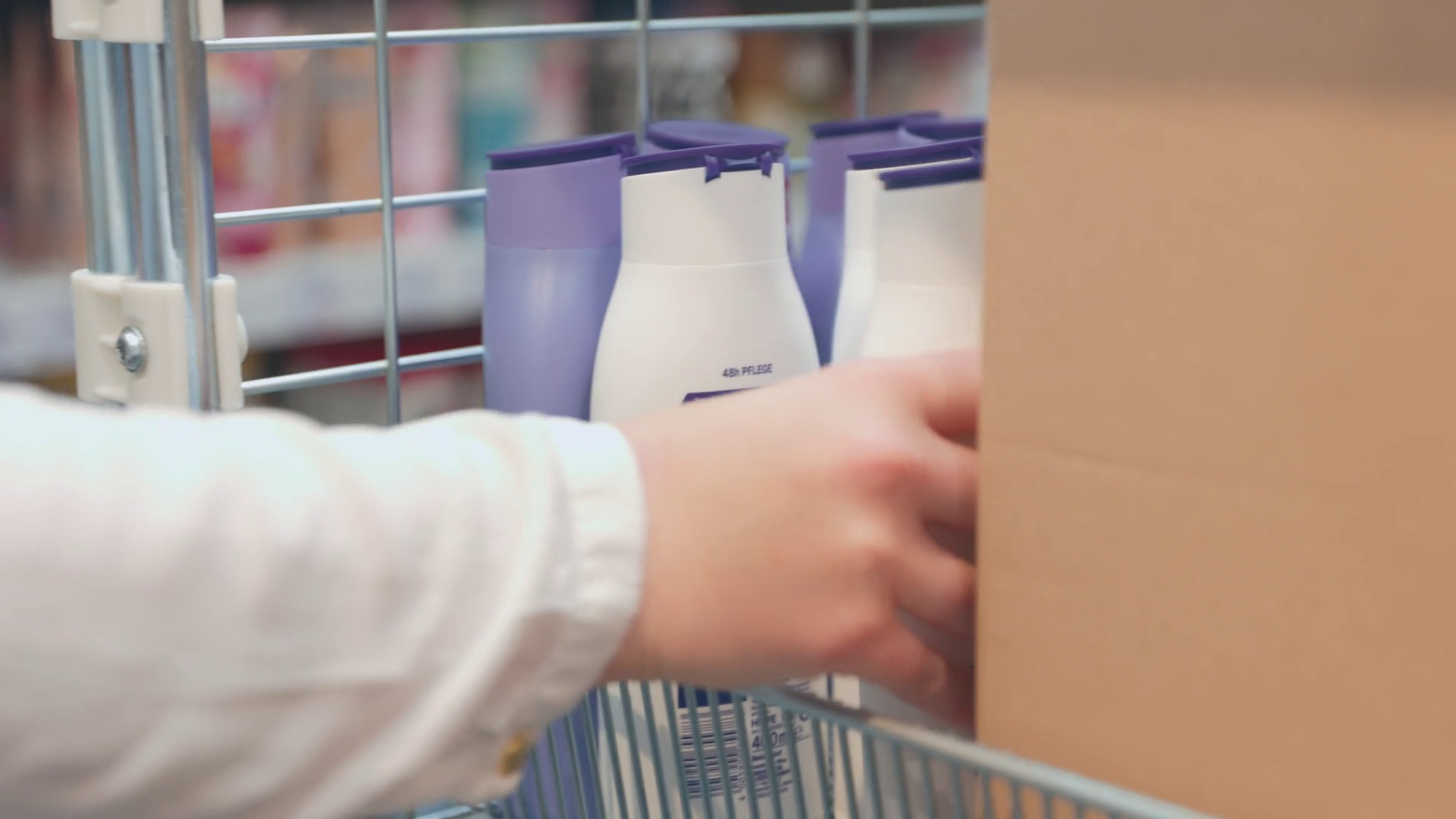}\hfill
\includegraphics[width=0.32\columnwidth]{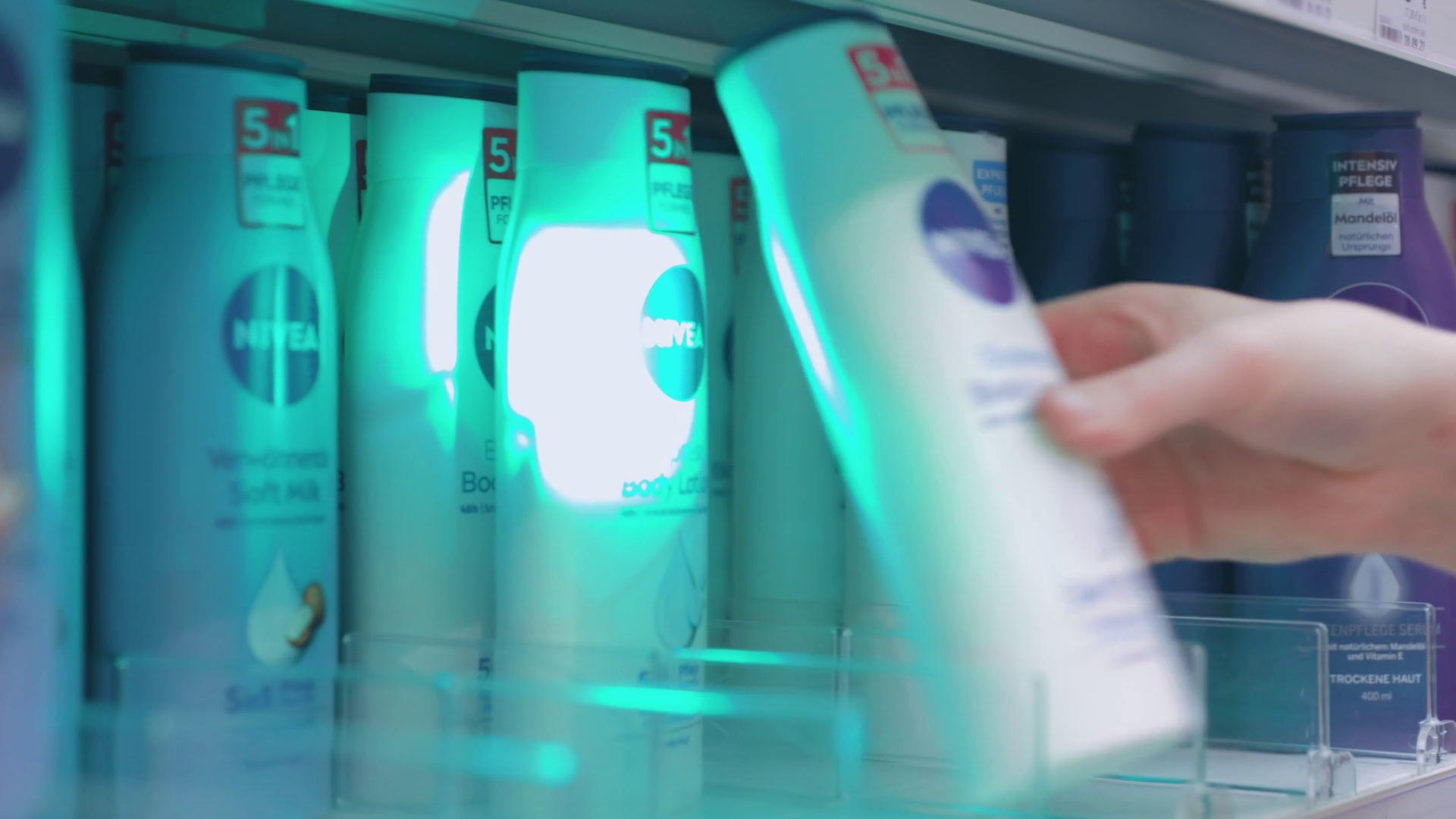}
\caption{Support replenishment by directing the store employee to the target location of the product.}
\label{fig:use_case_2}
\end{subfigure}\vspace{0.3cm} \\

\begin{subfigure}[t]{\textwidth}
\centering
\includegraphics[width=0.32\columnwidth]{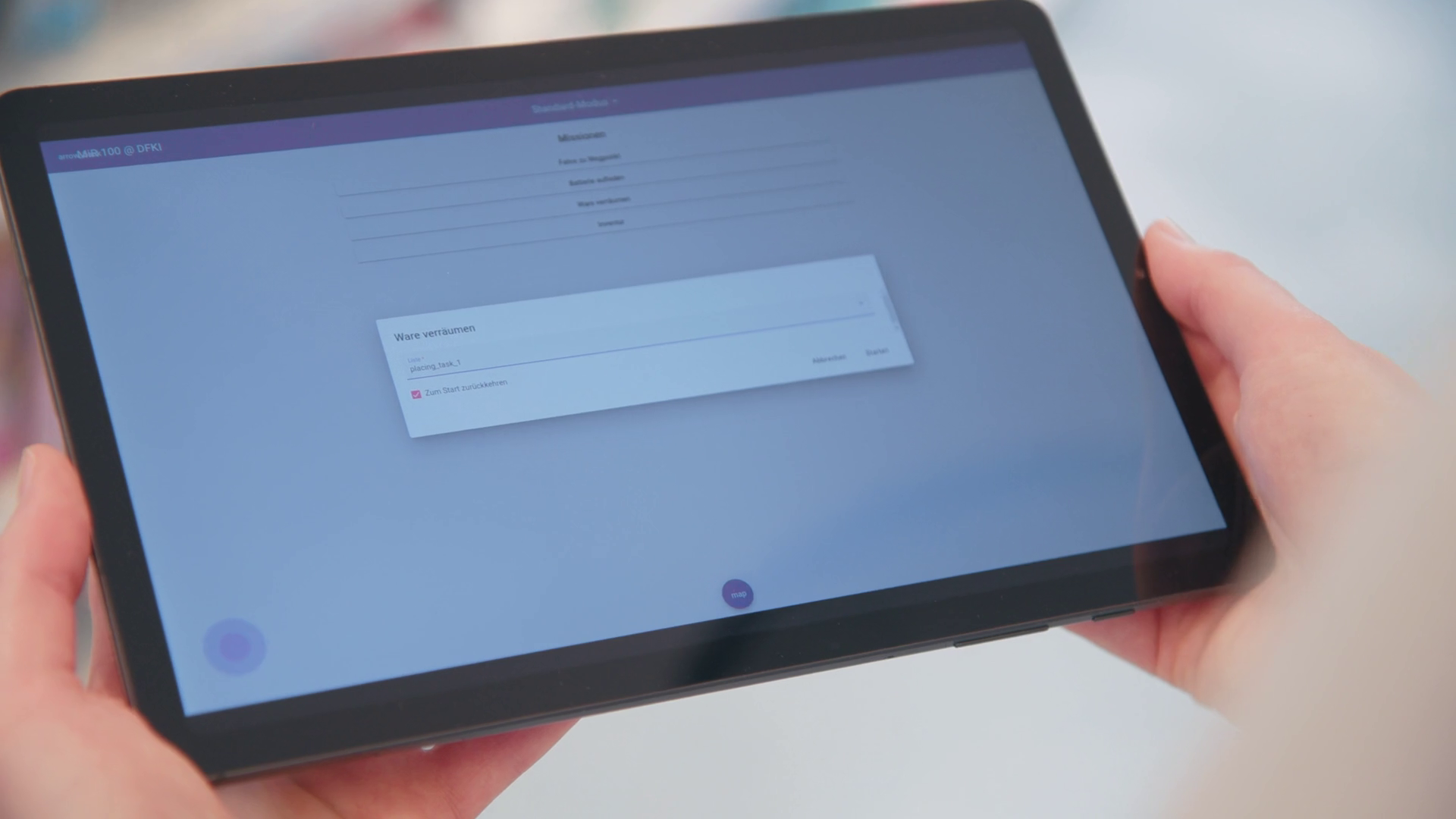}\hfill
\includegraphics[width=0.32\columnwidth]{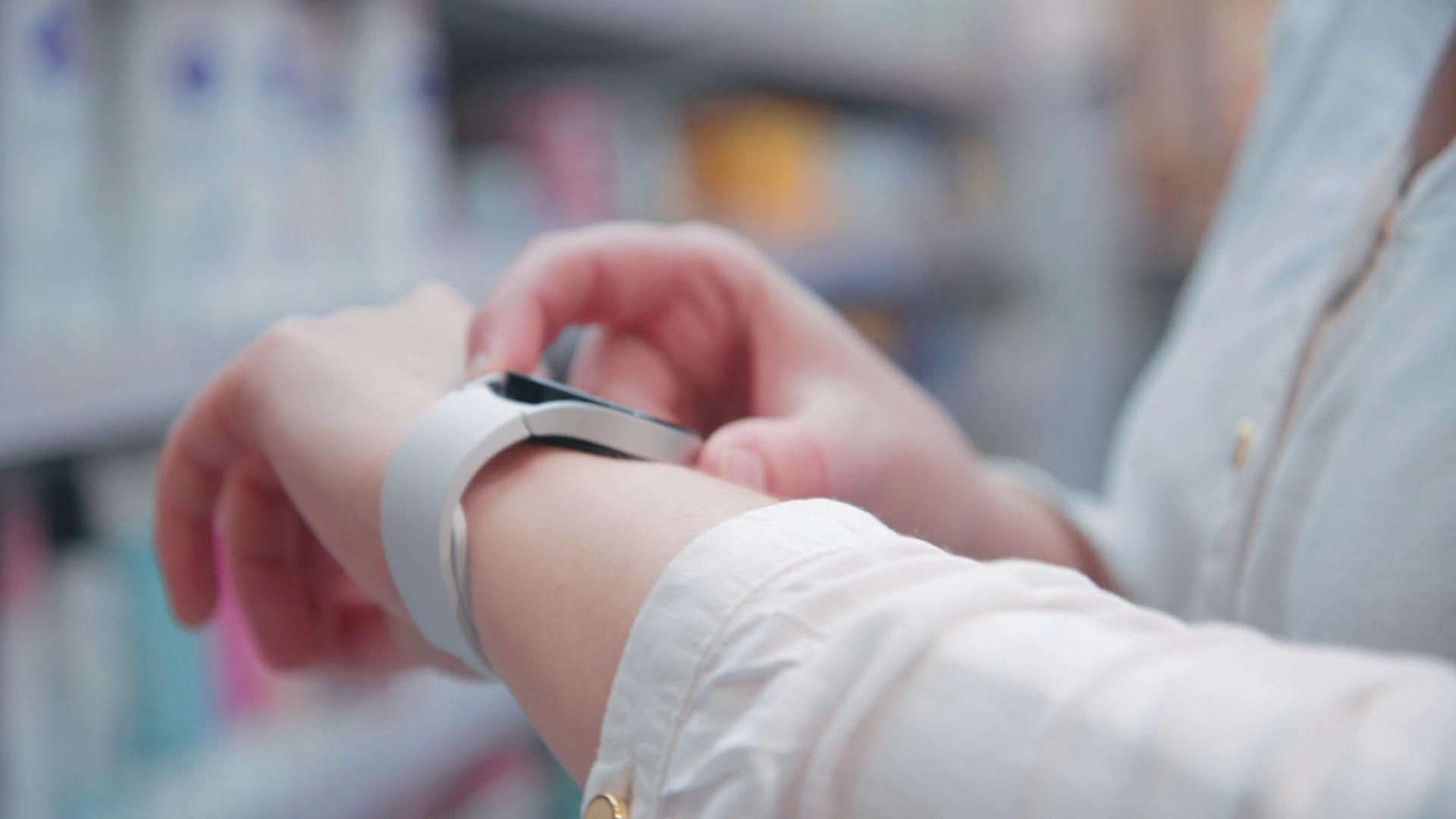}\hfill
\includegraphics[width=0.32\columnwidth]{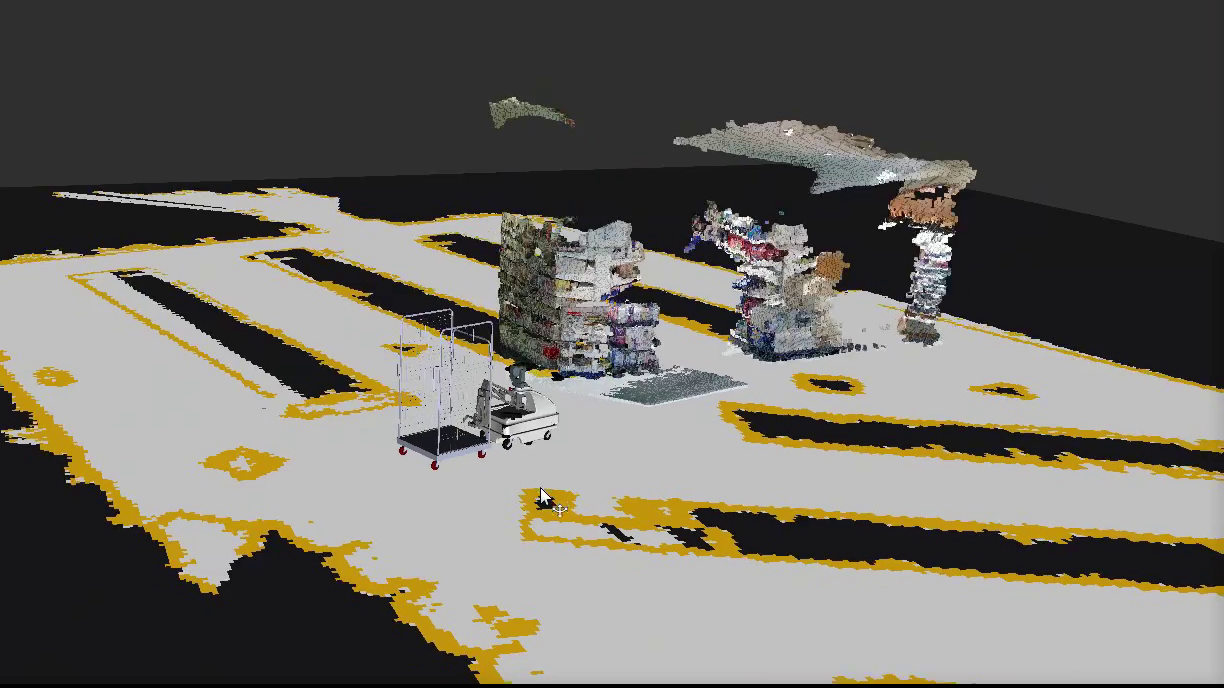}
\caption{User interfaces, GUI (\textit{left}), Smartwatch (\textit{middle}), Visualization of \RobotName{}'s sensor data (\textit{right}).}
\label{fig:use_case_3}
\end{subfigure}
\caption{Screenshots from a video taken during evaluation in a retail store.}
\label{fig:use_case}
\end{figure}

We evaluate the shelf replenishment application in a drugstore. \RobotName{} starts in the charging station.  We load different products from the store on the cart and manually pass the list of products to the robot. If the store employee selects the replenishment mission on the graphical user interface, \RobotName{} picks up  the cart with the products. Using the information in the semantic digital twin, the robot can infer the product locations, and, from the known store geometry, calculate feasible unloading points in front of the shelves where the products are located. Feasible means here that, firstly, the points have to be reachable for the robot navigation system. Second, if multiple products are located nearby in the same shelf they can be replenished using a single unloading point. And, third, the unloading point must be selected such that the robot does not block the target location in the shelf for the store employee. At every unloading point, the robot contacts the store employee, which receives a note on a smartwatch or tablet to start replenishment. If the task is finished, the employee must confirm this on the graphical user interface, and the robot will move to the next unloading point. This process is repeated until all products are unloaded from the cart. Afterwards, the robot will return to the charging station. During the entire process, \RobotName{} perceives the environment through its onboard sensors, detects and classifies obstacles, and stores permanent obstacles in the digital twin. Figure~\ref{fig:use_case} shows screenshots from a representative video demonstrating the use case.

%% file: sections/conclusion.tex
In this paper, we present the \RobotName{} service robotic system and its integration with the K4R platform, a cloud computing solution that enables AI and robotics applications for retail. By connecting with the K4R platform, \RobotName{}'s capabilities in perception, navigation, and mission planning are enhanced. We demonstrate that \RobotName{} is able to detect and classify unknown obstacles, navigate through the narrow aisles of a retail store, and plan and execute missions that assist the store employee in replenishing the shelves.

The potential of AI solutions and autonomous robotics in retail is huge. However, today the barriers to entry for retailers in such solutions are still quite high. For example, setting up a robotic system to support store employees requires a lot of expert knowledge and customized, expensive hardware. The idea of the K4R platform is to reduce these barriers to entry by providing retailers with the infrastructure and general-purpose AI functionalities. By centralizing AI approaches such as planning, reasoning, or machine learning in the K4R platform, it is possible to integrate commercially available robotic systems such as \RobotName{} into complex AI applications or even orchestrate entire fleets of AGVs. In this context, a possible extension of our work is to simultaneously use multiple robots for store intralogistics, for example by implementing a similar approach as in~\cite{Miranda2018} or scheduling methods as described in \cite{maas2023experimental}. Moreover, integrating different sensor sources, e.g., cameras to monitor the flow of customers, into the planning system can improve the reliability and speed of autonomous navigation in crowded stores. Finally, we would like to evaluate the proposed solution in a large-scale subject study (i.e., with store employees) to obtain realistic statements on usability and feasibility.
